\documentclass[journal,twoside,web]{ieeecolor}
\usepackage{graphicx}
\usepackage{CJKutf8}
\usepackage{amsmath}
\usepackage{amssymb}
\usepackage{booktabs}
\usepackage{multicol}
\usepackage{bbding}
\usepackage{makecell}
\usepackage{multirow}
\usepackage{float}
\usepackage{color, colortbl}
\usepackage{color}
\usepackage[dvipsnames]{xcolor}
\usepackage{listings}
\usepackage{courier}
\definecolor{codegreen}{rgb}{0,0.6,0}
\definecolor{codegray}{rgb}{0.5,0.5,0.5}
\definecolor{codepurple}{rgb}{0.58,0,0.82}
\definecolor{backcolour}{rgb}{1,1,1}
\lstdefinestyle{mystyle}{
    backgroundcolor=\color{backcolour},   
    commentstyle=\color{codegreen},
    keywordstyle=\color{magenta},
    numberstyle=\tiny\color{codegray},
    stringstyle=\color{codepurple},
    basicstyle=\ttfamily\footnotesize,
    breakatwhitespace=false,         
    breaklines=true,                 
    captionpos=b,                    
    keepspaces=true,                 
    numbers=none,                    
    numbersep=5pt,                  
    showspaces=false,                
    showstringspaces=false,
    showtabs=false,                  
    tabsize=2
}
\usepackage[lined,boxed,linesnumbered,ruled]{algorithm2e}
\usepackage{algorithm2e}

\newcommand{\tred}[1]{\textcolor{red}{#1}}
\newcommand{\tgrn}[1]{\textcolor{ForestGreen}{#1}}

\newcommand{\tpurp}[1]{\textcolor{DarkOrchid}{#1}}
\usepackage[pagebackref,breaklinks,colorlinks]{hyperref}
\definecolor{highlight}{rgb}{0, 0, 0}
\usepackage{arXiv-tmi}
\usepackage{cite}
\usepackage{amsmath,amssymb,amsfonts}
\usepackage{algorithmic}
\usepackage{graphicx}
\usepackage{textcomp}
\begin{document}
\title{ChatCAD+: Towards a Universal and Reliable  Interactive CAD using LLMs}
\author{Zihao Zhao, Sheng Wang, Jinchen Gu, Yitao Zhu, \\
Lanzhuju Mei, Zixu Zhuang, Zhiming Cui, Qian Wang, Dinggang Shen, \IEEEmembership{Fellow, IEEE}
\thanks{Zihao Zhao, Sheng Wang, Jinchen Gu and Yitao Zhu contributed equally and are listed as first authors.}
\thanks{Corresponding authors: Qian Wang and Dinggang Shen.}
\thanks{Zihao Zhao, Jinchen Gu, Yitao Zhu, Lanzhuju Mei, and Zhiming Cui are with the School of Biomedical Engineering, ShanghaiTech University, Shanghai 201210, China (email: zihaozhao10@gmail.com, \{gujch12022,zhuyt, meilzhj, cuizm\}@shanghaitech.edu.cn)}
\thanks{Sheng Wang and Zixu Zhuang are with the School of Biomedical Engineering, Shanghai Jiao Tong University, Shanghai 200030, China,  Shanghai United Imaging Intelligence Co., Ltd., Shanghai 200230, China, and also ShanghaiTech University, Shanghai 201210, China. (e-mail:  \{wsheng, zixuzhuang\}@sjtu.edu.cn)}
\thanks{Qian Wang and Dinggang Shen are with the School of Biomedical Engineering \& State Key Laboratory of Advanced Medical Materials and Devices, ShanghaiTech University, Shanghai 201210, China, and Shanghai Clinical Research and Trial Center, Shanghai 201210, China. Dinggang Shen is also with Shanghai United Imaging Intelligence Co. Ltd., Shanghai 200230, China. (e-mail: \{qianwang, dgshen\}@shanghaitech.edu.cn)
}
}

\maketitle

\begin{abstract}
The integration of Computer-Aided Diagnosis (CAD) with Large Language Models (LLMs) presents a promising frontier in clinical applications, notably in automating diagnostic processes akin to those performed by radiologists and providing consultations similar to a virtual family doctor. Despite the promising potential of this integration, current works face at least two limitations:
\textcolor{highlight}{
(1) From the perspective of a radiologist, existing studies typically have a restricted scope of applicable imaging domains, failing to meet the diagnostic needs of different patients. Also, the insufficient diagnostic capability of LLMs further undermine the quality and reliability of the generated medical reports. (2) Current LLMs lack the requisite depth in medical expertise, rendering them less effective as virtual family doctors due to the potential unreliability of the advice provided during patient consultations.}
To address these limitations, we introduce ChatCAD+, to be universal and reliable. 
\textcolor{highlight}{
Specifically, it is featured by two main modules: (1) Reliable Report Generation and (2) Reliable Interaction. The Reliable Report Generation module is capable of interpreting medical images from diverse domains and generate high-quality medical reports via our proposed hierarchical in-context learning. Concurrently, the interaction module leverages up-to-date information from reputable medical websites to provide reliable medical advice. Together, these designed modules synergize to closely align with the expertise of human medical professionals, offering enhanced consistency and reliability for interpretation and advice.} The source code is available at \href{https://github.com/zhaozh10/ChatCAD}{GitHub}.
\end{abstract}

\section{Introduction}
\label{sec:introduction}
\IEEEPARstart{L}{arge} Language Models (LLMs) have emerged as promising tools in various domains. It refers to advanced artificial intelligence that has been extensively trained on text data.
Drawing on the combined power of sophisticated deep learning techniques, large-scale datasets, and increased model sizes, LLMs demonstrate extraordinary capabilities in understanding and generating human-like text. This is substantiated by significant projects like ChatGPT~\cite{OpenAI2023ChatGPT} and LLaMA~\cite{touvron2023llama}. 
These techniques are highly suitable for a wide range of scenarios, such as customer service, marketing, education, and healthcare consultation. Notably, ChatGPT has passed part of the US medical licensing exams, highlighting the potential of LLMs in the medical domain~\cite{qiu2023large,xue2023potential}. 

\begin{figure}[tbp]
    \centering
    \includegraphics[width=0.45\textwidth]{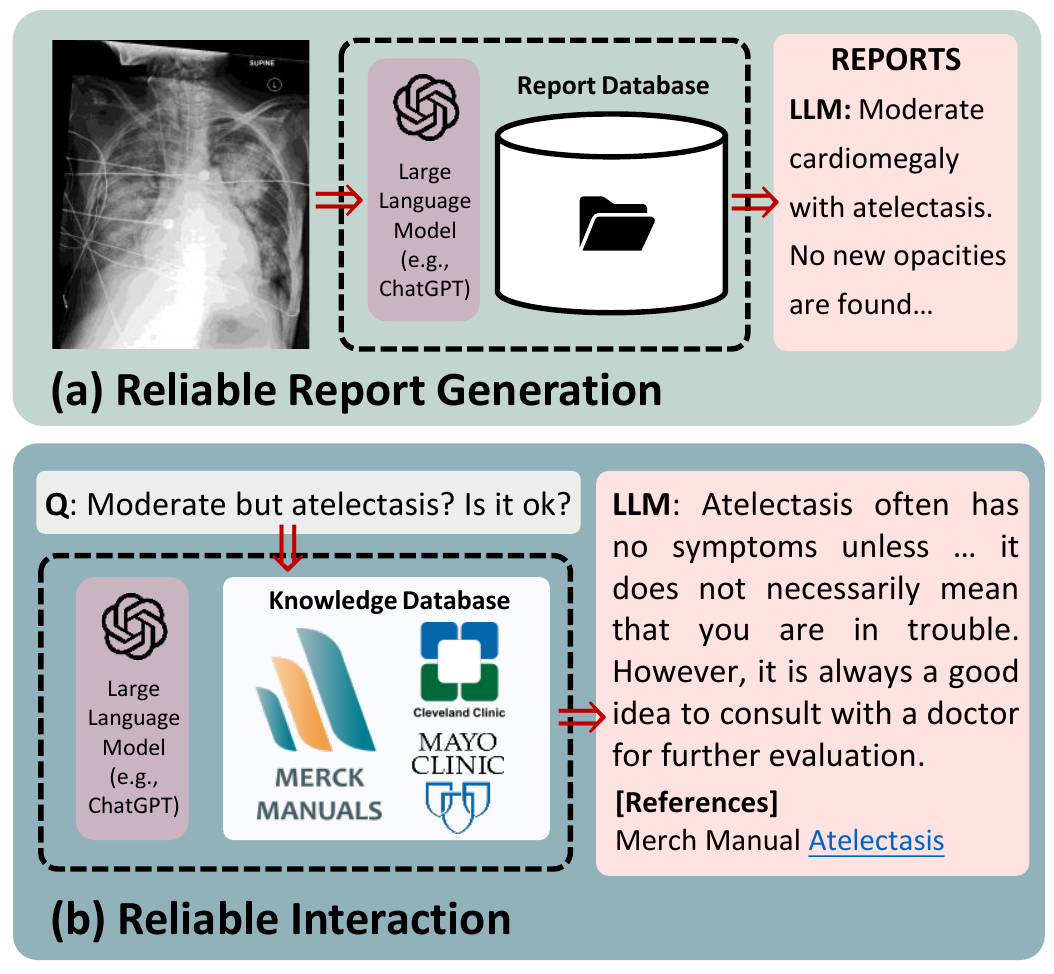}
    \caption{\textcolor{highlight}{Overview of our proposed ChatCAD+ system. (a) For patients seeking a diagnosis, ChatCAD+ generates reliable medical reports based on the input medical image(s) by referring to local report database. (b) Additionally, for any inquiry from patients, ChatCAD+ retrieves related knowledge from online database and lets large language model generate reliable response.}}
    \label{fig:teaser}
\end{figure}

So far, several studies have been conducted to integrate LLMs into Computer-Assisted Diagnosis (CAD) systems of medical images.
Conventional CAD typically operates following pure computer vision~\cite{cui2022fully,wang2022follow} or vision-language paradigms~\cite{chen-acl-2021-r2gencmn,chen2023visual}.
LLMs have shown ability to effectively interpret findings from medical images, thus mitigating limitations of only visual interpretation. 
For example, in our pilot study, we leverage the intermediate results obtained from image CAD networks and then utilize LLMs to generate  final diagnostic reports~\cite{wang2023chatcad}.

Although some efforts have been made to combine CAD networks and LLMs~\cite{wang2023chatcad,milecki2023medimp,niu2023ct}, it should be noted that these studies are limited in their scopes, which often focus on specific image domains. That is, such a system may only support a single image modality, organ, or application (such as chest X-ray), which greatly limits generalizability in the real clinical workflow. 
The primary reason for this limitation comes from notable topologic and semantic variations observed among medical image data, which present distinct challenges when attempting to encode various images with a unified model. 
Furthermore, although LLMs have demonstrated the capability to articulate medical concepts and clinical interpretations~\cite{ueda2023diagnostic,liu2023utility},
 there still exists a significant gap between LLMs and radiologists. This gap undoubtedly undermines patient confidence in a LLM-generated report. Consequently, these issues hinder practical utilization of LLMs in automated generation of medical reports.
In addition, it is also articulable that the integration of LLMs and CADs may lead to medical dialogue systems~\cite{yunxiang2023chatdoctor,xiong2023doctorglm,wang2023huatuo}. 
In this way, patients will be able to interact through LLMs and acquire more medical advice and explanation, while this functionality is often missing in conventional CAD systems. 
However, existing studies show that the general LLMs typically produce medical advice based solely on their encoded knowledge, without considering specific knowledge in the medical domain~\cite{xiong2023doctorglm}. 
As a result, patients may receive unreliable responses, dwarfing trust in CADs and thereby hindering the use of such systems in real medical scenarios. 

\begin{figure*}[t]
    \centering
    \includegraphics[width=0.95\textwidth]{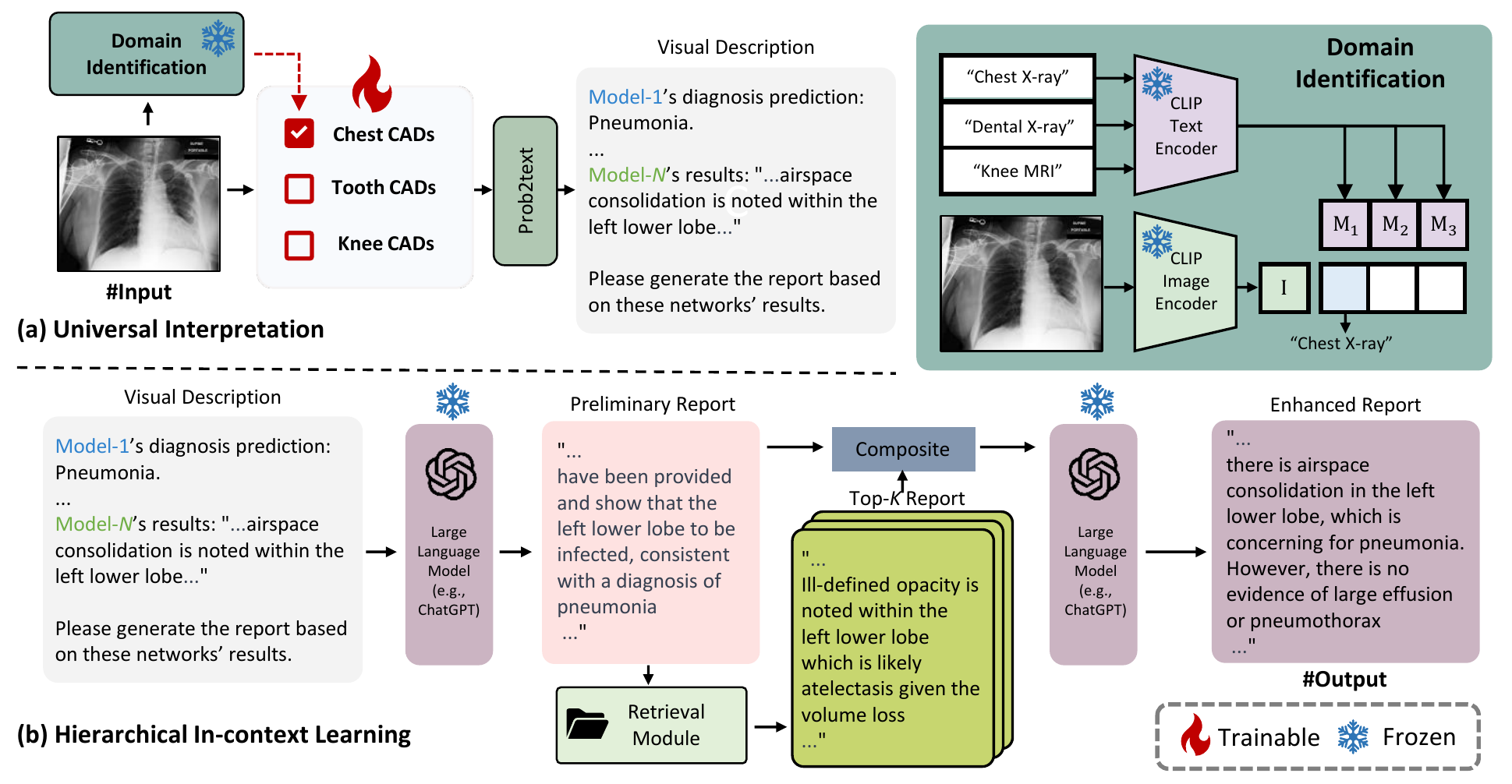}
    \caption{\textbf{Overview of the reliable report generation.} (a) Universal interpretation: To enhance the vision capability of LLMs, multiple CAD models are incorporated. The numerical results obtained from these models are transformed into visual descriptions following the rule of \textbf{prob2text}. (b) 
    Hierarchical in-context learning:
    The LLM initially generates a preliminary report based on the visual description, which is then enhanced through in-context learning using retrieved semantically similar reports. LLM and CLIP models are kept frozen, while CAD models are denoted as trainable as they can be continually trained and updated without interfering with other components.}
    \label{fig:overview}
\end{figure*}

In order to tackle the challenges mentioned above, we propose ChatCAD+ in this paper. 
And our contributions are made in the following aspects.
\textbf{
(1) Universal image interpretation.} Due to the difficulty in obtaining a unified CAD network tackling various images currently, ChatCAD+ incorporates a domain identification module to work with a variety of CAD models (c.f. Fig. \ref{fig:overview}(a)). ChatCAD+ can adaptively select a corresponding model given the input medical image. The tentative output of the CAD network is converted into text description to reflect image features, making it applicable for diagnostic reporting subsequently.  
\textcolor{highlight}{
\textbf{
(2) Hierarchical in-context learning for enhanced report generation. }}
Top-\textit{k} reports that are semantically similar to the LLM-generated report are retrieved from a clinic database via the proposed retrieval module (c.f. Fig. \ref{fig:retrieval}. The retrieved \textit{k} reports then serve as in-context examples to refine the LLM-generated report.
\textbf{
(3) Knowledge-based reliable interaction.} As illustrated in Fig.~\ref{fig:teaser}(b), ChatCAD+ does not directly provide medical advice. Instead, it first seeks help via our proposed knowledge retrieval module for obtaining relevant knowledge from professional sources, e.g. Merck Manuals, Mayo Clinic, and Cleveland Clinic. Then, the LLM considers the retrieved knowledge as a reference to provide reliable medical advice. 

In summary, our ChatCAD+ for the first time builds a universal and reliable medical dialogue system. The improved quality of answers and diagnostic reports of the chatbot reveals potential of LLMs in interactive medical consultation.



\section{Related Works}
\subsection{Large Language Models in Healthcare}

Following the impressive potential demonstrated by ChatGPT with its 100B-scale parameters, researchers have further expanded the application of language models in healthcare unprecedentedly, yielding highly promising results. Med-PaLM~\cite{singhal2022large} was developed using meticulously curated biomedical corpora and human feedback, showcasing its potential with notable achievements, including a remarkable 67.6\% accuracy on the MedQA exam. Notably, ChatGPT, which did not receive specific medical training, successfully passed all three parts of the USMLE and achieved an overall accuracy of over 50\% across all exams, with several exams surpassing 60\% accuracy~\cite{kung2022performance}. \textcolor{highlight}{ChatDoctor~\cite{yunxiang2023chatdoctor}, fine-tuned on the LLaMA model using synthetic clinical QA data generated by ChatGPT, also incorporated medical domain knowledge to provide reliable responses.
However, our ChatCAD+ distinguishes its reliable interaction from ChatDoctor in several aspects. First, ChatDoctor relies on Wikipedia, which is editable for all users and may compromise the rigor and reliability required in the medical domain. Second, ChatDoctor simply divides the knowledge of each medical topic into equal sections and relies on rule-based ranking to select relevant information. In contrast, our reliable interaction leverages the Merck Manual~\cite{porter2011merck}, a comprehensive and trusted medical knowledge resource. It can organize medical knowledge into a hierarchical structure, thus enabling us to fully exploit the reasoning capabilities of the LLM in recursively retrieving the most pertinent and valuable knowledge.}

\subsection{Multi-modality Large Models}
In present times, multi-modality large models are typically constructed based on pre-trained models rather than training them from scratch. Frozen~\cite{tsimpoukelli2021multimodal} performed fine-tuning on an image encoder, utilizing its outputs as soft prompts for the language model. Flamingo~\cite{alayrac2022flamingo} introduced cross-attention layers into the LLM to incorporate visual features, exclusively pre-training these new layers. BLIP-2~\cite{li2023blip} leveraged both frozen image encoders and frozen LLMs, connecting them through their proposed Q-Former.
\textcolor{highlight}{Following the architecture of BLIP-2, XrayGPT~\cite{thawkar2023xraygpt} was specifically developed on chest X-ray. While the training cost for XrayGPT is demanding, our proposed ChatCAD+ can seamlessly integrate existing pre-trained CAD models without requiring further training. This highlights the inherent advantage of ChatCAD+ in continual learning scenarios, where additional CAD models can be effortlessly incorporated.}

In more recent studies, researchers have taken a different approach, i.e., designing prompts to allow LLMs to utilize vision models for avoiding training altogether. For instance, Visual-ChatGPT~\cite{wu2023visual} established a connection between ChatGPT and a series of visual foundation models, enabling exchange of images during conversations. ChatCAD~\cite{wang2023chatcad} linked existing CAD models with LLMs to enhance diagnostic accuracy and improve patient care.
\textcolor{highlight}{Following a similar} \textcolor{highlight}{paradigm as ChatCAD~\cite{wang2023chatcad}, Visual Med-Alpaca~\cite{visualmedAlpaca} incorporated multiple vision models to generate text descriptions for medical images.} \textcolor{highlight}{However, when compared to our study,
it is limited to a single medical imaging domain and also lacks reliability in both report generation and interactive question answering.}


\section{Method}

The proposed ChatCAD+ is a multi-modality system capable of processing both image and text inputs (as illustrated in Fig.~\ref{fig:teaser}) through reliable report generation and reliable interaction, respectively.
It is important to note that, in this paper, the term ``modality" refers to language and image, which differs from the commonly used concept of medical image modality.
The reliable report generation treats medical image as input, identifies its domain, and provides enhanced medical reports by referencing a local report database. 
Simultaneously, for the user's text query, the reliable interaction will retrieve clinically-sound knowledge from a professional knowledge database, thereby serving as a reliable reference for the LLM (e.g. ChatGPT) to enhance the reliability its response.

\subsection{Reliable Report Generation}
\begin{figure}[t]
    \centering
    \includegraphics[width=0.5\textwidth]{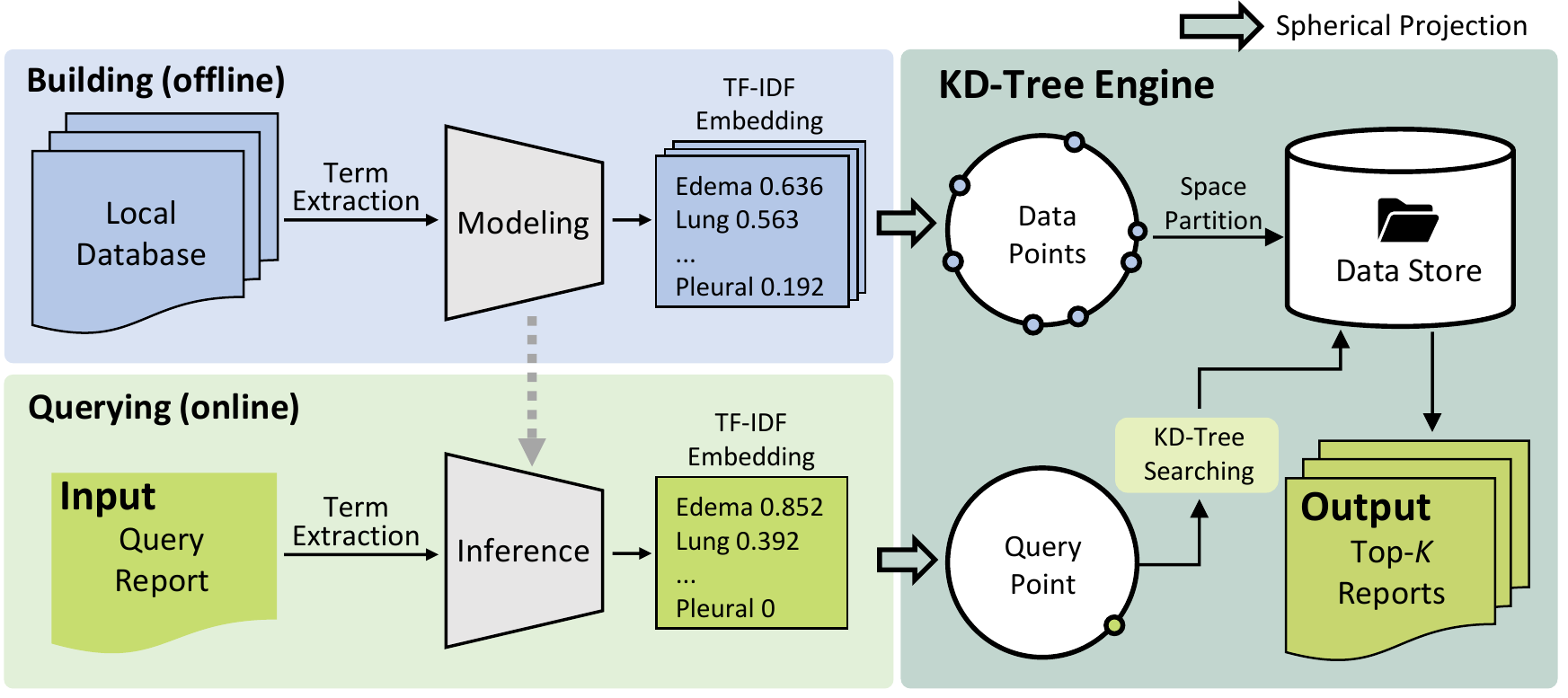}
    \caption{\textcolor{highlight}{\textbf{The illustration of the retrieval module within reliable report generation.} It adopts the TF-IDF algorithm to preserve the semantics of each report and converts it into a latent embedding during offline modeling and online inference.
    To facilitate highly efficient retrieval, we perform spherical projection on all TF-IDF embeddings, whether during building or querying. In this manner, we can utilize the KD-Tree structure to store these data and implement retrieval with a low time complexity.}}
    \label{fig:retrieval}
\end{figure}

\textcolor{highlight}{
This section outlines our proposed reliable report generation, as depicted in Fig.~\ref{fig:overview}, which encompasses two primary stages: universal interpretation and hierarchical in-context generation. 
Section~\ref{sec:universal} illustrates the paradigm to interpret medical images from diverse domains into descriptions that can be understood by LLMs. The resulting visual description then serves as the input for subsequent hierarchical in-context learning.
Following this, Section~\ref{sec:hierarchical} elucidates hierarchical in-context learning, within which the retrieval module (in Fig.~\ref{fig:retrieval}) acts as key component.}

\subsubsection{Universal Interpretation}\label{sec:universal}
\textcolor{highlight}{
The process of universal interpretation can be divided into three steps. Initially, the proposed {domain identification} module is employed to determine the specific domain of the medical image. Subsequently, the corresponding domain-specific model is activated to interpret the image. Finally, the interpretation result is converted into a text prompt for describing visual information via the rule-based {prob2text} module for further processing.}

\begin{table}[tbp]
  \centering
  \caption{The illustration of prob2text. P3 is the default setting.}
    \begin{tabular}{rl}
    \multicolumn{2}{l}{CAD Model$\to$(\textbf{disease}: \textbf{prob})} \\
    \midrule
    \multicolumn{1}{l}{P1 (direct)} & ``\{\textbf{disease}\} score: \{\textbf{prob}\}'' \\
    \midrule
    \multicolumn{1}{l}{\multirow{2}[2]{*}{P2 (simplistic)}} & \textbf{prob}$\in$[0, 0.5): ``No Finding'' \\
          & \textbf{prob}$\in$[0.5, 1]: ``The prediction is \{\textbf{disease}\}'' \\
    \midrule
    \multicolumn{1}{l}{\multirow{4}[2]{*}{P3 (illustrative)}} & \textbf{prob}$\in$[0, 0.2): ``No sign of \{\textbf{disease}\}'' \\
          & \textbf{prob}$\in$[0.2, 0.5): ``Small possibility of \{\textbf{disease}\}'' \\
          & \textbf{prob}$\in$[0.5, 0.9): ``Patient is likely to have \{\textbf{disease}\}'' \\
          & \textbf{prob}$\in$[0.9, 1]: ``Definitely have \{\textbf{disease}\}'' \\
    \bottomrule
    \end{tabular}%
  \label{tab:prompt}%
\end{table}%

Domain identification chooses CAD models to interpret the input medical image, which is fundamental for subsequent operations of ChatCAD+. To achieve this, we employ a method of visual-language contrast that computes the cosine similarity between the input image and textual representations of various potential domains. This approach takes advantage of language's ability in densely describing the features of a particular type of image, as well as its ease of extensibility.
In particular, we employ the pre-trained BiomedCLIP model~\cite{zhang2023large} to encode both the medical image and the text associated with domains of interest. 
In this study, we demonstrate upon three domains: chest X-ray, dental X-ray, and knee MRI. The workflow is depicted in the upper right portion of Fig.~\ref{fig:overview}. Assuming there are three domains ${\textrm{D}_1, \textrm{D}_2, \textrm{D}_3}$, along with their textual representations ${\textrm{M}_1, \textrm{M}_2, \textrm{M}_3}$, and also a visual representation denoted as $\textrm{I}$ for the input image, we define
\begin{equation}
    \textrm{D}_\textrm{pred}=\operatorname{argmax}_{i\in\{1, 2, 3\}} \frac{\textrm{I} \ \textrm{M}_i}{\left\Vert \textrm{I}\right\Vert\left\Vert \textrm{M}_i\right\Vert},
    \label{clip}
\end{equation}
where $\textrm{D}_\textrm{pred}$ denotes the prediction of the medical image domain. The module thereby can call the domain-specific CAD model to analyze visual information given $\textrm{D}_\textrm{pred}$.

Since most CAD models generate outputs that can hardly be understood by language models, further processing is needed to bridge this gap.
For example, an image diagnosis model typically outputs tensors representing the likelihood of certain clinical findings. To establish a link between image and text, these tensors are transformed into textual descriptions according to diagnostic-related rules, which is denoted as {prob2text} in Fig.~\ref{fig:overview}(a). 
The prob2text module has been designed to present clinically relevant information in a manner that is more easily interpretable by LLMs. 
The details of the prompt design are illustrated in Table~\ref{tab:prompt}. 
Using chest X-ray as an example, we follow the three types (P1-P3) of prompt designs in~\cite{wang2023chatcad} and adopt \textbf{P3 (illustrative)} as the recommended setting in this study. Concretely, it employs a grading system that maps the numerical scores into clinically illustrative description of disease severity. 
The scores are divided into four levels based on their magnitude: ``No sign'', ``Small possibility'', ``Likely'', and ``Definitely''. 
The corresponding texts are then used to describe the likelihood of different observations in chest X-ray, providing a concise and informative summary of the patient's condition. 
The prompt design for dental X-ray and knee MRI are in similar ways.
And other prompt designs, such as P1 and P2~\cite{wang2023chatcad}, will be discussed in experiments below. 

\subsubsection{Hierarchical In-context Learning}\label{sec:hierarchical}
\textcolor{highlight}{
The proposed strategy for hierarchical in-context learning, as shown in Fig.~\ref{fig:overview}(b), consists of (1) preliminary report generation and (2) retrieval-based in-context enhancement.}

\textcolor{highlight}{
In the first step, the LLM utilizes the visual description to generate a preliminary diagnostic report. This process is facilitated through prompts such as ``Write a report based on results from Network(s)." It is important to note that the generation of this preliminary report relies solely on the capability of the LLM and is still far from matching capability of human specialists. As a result, even powerful LLMs like ChatGPT still struggle to achieve satisfactory results.}

\textcolor{highlight}{
To address this concern, the next step involves retrieving the top-\textit{k} reports that share similar semantics with the preliminary report as in-context examples. The retrieval module is designed to preserve semantic information and ensure efficient implementation. Specifically, we employ TF-IDF \cite{sparck1972statistical} to model the semantics of each report, and the KD-Tree \cite{bentley1975multidimensional} data structure is utilized to facilitate high-speed retrieval.}
 \textcolor{highlight}{
    Fig.~\ref{fig:retrieval} illustrates the detailed pipeline of the proposed retrieval module. For simplification, we use chest X-ray in MIMIC-CXR as an example in this section. To better preserve disease-related information in the report, we focus on 17 medical terms related to thoracic diseases during the modeling of TF-IDF embeddings, which form a term set $\mathcal{T}$. Assuming the training set of MIMIC-CXR as $\mathcal{D}$, the TF-IDF score for each term $t \in \mathcal{T}$ in each document $d \in \mathcal{D}$ is computed as:
\begin{align}
    &\textrm{TF\text{-}IDF}(t,d)=\textrm{TF}(t,d)\cdot \textrm{IDF}(t)\notag\\
    &=\frac{| { \{w \in d: w=t\} } |}{|d|}\cdot\log{\frac{|\mathcal{D}|}{|\{d\in\mathcal{D}: t\in d\}|}},
    \end{align}}
    \textcolor{highlight}{
where $w$ indicates a word within the document $d$. In brief, TF$(t,d)$ represents the frequency of term $t$ occurring in $d$, and IDF$(t)$ denotes the inverse of the frequency that document $d$ occurs in the training set $\mathcal{D}$. Following this, we define the TF-IDF embedding (TIE) of a document $d$ as:
\begin{align}
    \textrm{TIE}(d)=\{{\textrm{TF-IDF}(t,d)| t\in\mathcal{T}}\}.
\end{align}}
\begin{figure*}[tbp]
    \centering
    \includegraphics[width=0.85\textwidth]{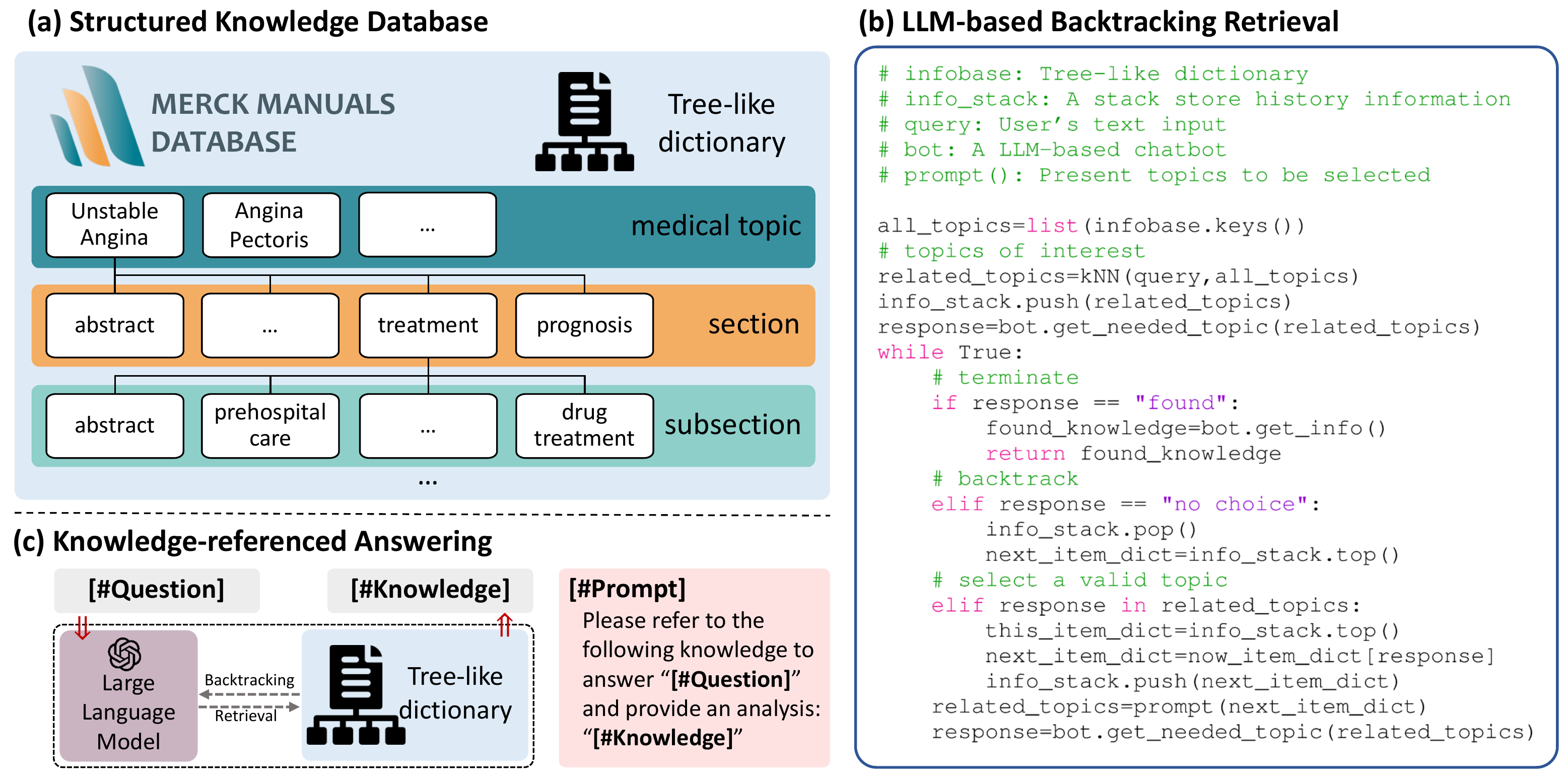}
    \caption{
    \textcolor{highlight}{Overview of the reliable interaction. (a) 
    The illustration of structured medical knowledge database, organized as a tree-like dictionary, where each medical topic has multiple sections while sections can be further divided into subsections. (b) A LLM-based knowledge retrieval method is proposed to search relevant knowledge in a backtrack manner. (c) The LLM is prompted to answer the question based on the retrieved knowledge.}}
    \label{fig:interaction_overview}
\end{figure*}

\textcolor{highlight}{
After calculating the TIEs for all $d \in \mathcal{D}$, we organize them using the KD-Tree \cite{bentley1975multidimensional} data structure, which enables fast online querying. The KD-Tree is a space-partitioning data structure that can implement \textit{k}-nearest-neighbor querying in $O(\log(n))$ time on average, making it suitable for information searching. However, since the implementation of KD-Tree relies on $L_2$ distance instead of cosine similarity, it cannot be directly applied to our TIEs. To address this issue, we project all TIEs onto the surface of a unit hypersphere, as shown in the right panel of Fig.~\ref{fig:retrieval}. After spherical projection, we have:
\begin{align}
    L_2({\frac{\vec{q}}{|\vec{q}|}},{\frac{\vec{v}}{|\vec{v}|}})=2r \cdot \sin({\frac{\theta}{2}}),
\end{align}
where $\vec{q}$ and $\vec{v}$ represent the TIE of the query report and a selected report, respectively. $r$ indicates the radius of the hypersphere (radius=1 in this study), and $\theta \in [0,\pi]$ stands for the angle between $\frac{\vec{q}}{|\vec{q}|}$ and $\frac{\vec{v}}{|\vec{v}|}$. The $L_2$ distance between two TIEs will monotonically increase as a function of the angle between them, which favors the application of the KD-Tree data structure for efficient retrieval.}

\textcolor{highlight}{
During online querying, the KD-Tree engine retrieves the top-\textit{k} (i.e., \textit{k}=3) reports that share the most} \textcolor{highlight}{similar semantics with the query report (i.e., preliminary report illustrated in Fig.~\ref{fig:overview}). It then asks the LLM to refine its preliminary report with the retrieved \textit{k} reports as references, and generate the enhanced report as the final output.}

\subsection{Reliable Interaction}
\textcolor{highlight}{
This section outlines the reliable interaction, which is responsible for answering the user's text query.
It serves a dual role within our proposed ChatCAD+. First, reliable interaction can function as a sequential module of reliable report generation to enhance interpretability and understandability of medical reports for patients. In this capacity, it takes the generated report from previous module as the input, and provides patients with explanations of involved clinical terms\footnote{Clinical terms within a radiology report can be automatically extracted via tools like RadGraph~\cite{jain2021radgraph}.}.
Second, it can also operate independently to answer clinical-related questions and provide reliable medical advice.}

The proposed ChatCAD+ offers reliable interaction via the construction of a professional knowledge database and LLM-based knowledge retrieval. In this study, we demonstrate using the Merck Manuals. 
The Merck Manuals are a series of healthcare reference books that provide evidence-based information on the diagnosis and treatment of diseases and medical conditions.

 The implementation of LLM-based knowledge retrieval leverages the Chain-of-Thought (CoT) technique, which is widely known for its ability to enhance the performance of LLM in problem-solving. CoT breaks down a problem into a series of sub-problems and then synthesizes the answers to these sub-problems progressively for solving the original problem. Humans typically do not read the entire knowledge base but rather skim through topic titles to find what they are looking for. Inspired by this, we have designed a prompt system that automatically guides the LLM to execute such searches. Correspondingly, we have structured the Merck Manuals database as a hierarchical dictionary, with topic titles of different levels serving as keys, as shown in Fig~\ref{fig:interaction_overview}(a).
Fig.~\ref{fig:interaction_overview}(b) demonstrates the proposed knowledge retrieval methods. Initially, we provide the LLM with only titles of five related medical topics in the database, and then ask the LLM to select the most relevant topic to begin the retrieval. Once the LLM has made its choice, we provide it with the content of ``abstract'' section  and present names of all other sections subject to this topic. 
Given that medical topics often exhibit hierarchical organization across three or four tiers, we repeat this process iteratively, empowering the LLM to efficiently navigate through all tiers of a given medical topic. Upon identifying relevant medical knowledge, the LLM is prompted to return the retrieved knowledge, terminating the process. Otherwise, in the case that the LLM finds none of the provided information as relevant, our algorithm would backtrack to the parent tier, which makes it a kind of depth-first search~\cite{cormen2022introduction}. Finally, the LLM is prompted to provide reliable response based on the retrieved knowledge (in Fig.~\ref{fig:interaction_overview}(c)).

\begin{table*}[tbp]\small
  \centering
  \caption{Statistics of the composed dataset for the evaluation of domain identification.}
  \resizebox{.95\textwidth}{!}{
    \begin{tabular}{llllllllll}
    \toprule
     & chest X-ray   & dental X-ray     & knee X-ray    & mammography     & chest CT     & fundus     & endoscopy     & dermoscopy& blood cell \\
    \midrule
    number& 400 & 360 & 800 & 360      & 320     & 120     & 200 & 360 &200 \\
    source & MedFMC~\cite{wang2023real}& Tufts Dental~\cite{panetta2021tufts}  & OAI~\cite{chen2019fully}  & INbreast~\cite{InsMoreira2012INbreastTA} &   COVID-CT~\cite{zhao2020COVID-CT-Dataset}   &  DIARETDB1~\cite{kauppi2007diaretdb1}     & MedFMC~\cite{wang2023real}     &ISIC2016~\cite{gutman2016skin}       & PBC~\cite{acevedo2020dataset} \\
    \bottomrule
    \end{tabular}
    }%
  \label{tab:uni_stat}%
\end{table*}%

\section{Experimental Results}\label{exp}
\textcolor{highlight}{
To comprehensively assess our proposed ChatCAD+, we conducted evaluations across three crucial aspects that influence its effectiveness in real-world scenarios: (1) the capability to handle medical images from different domains, (2) report generation quality, and (3) the efficacy in clinical question answering. This section commences with an introduction to the datasets used and implementation details of ChatCAD+. Then, we delve into the evaluation of these aspects in turn.}

\begin{figure}[tbp]
    \centering
    \includegraphics[width=0.45\textwidth]{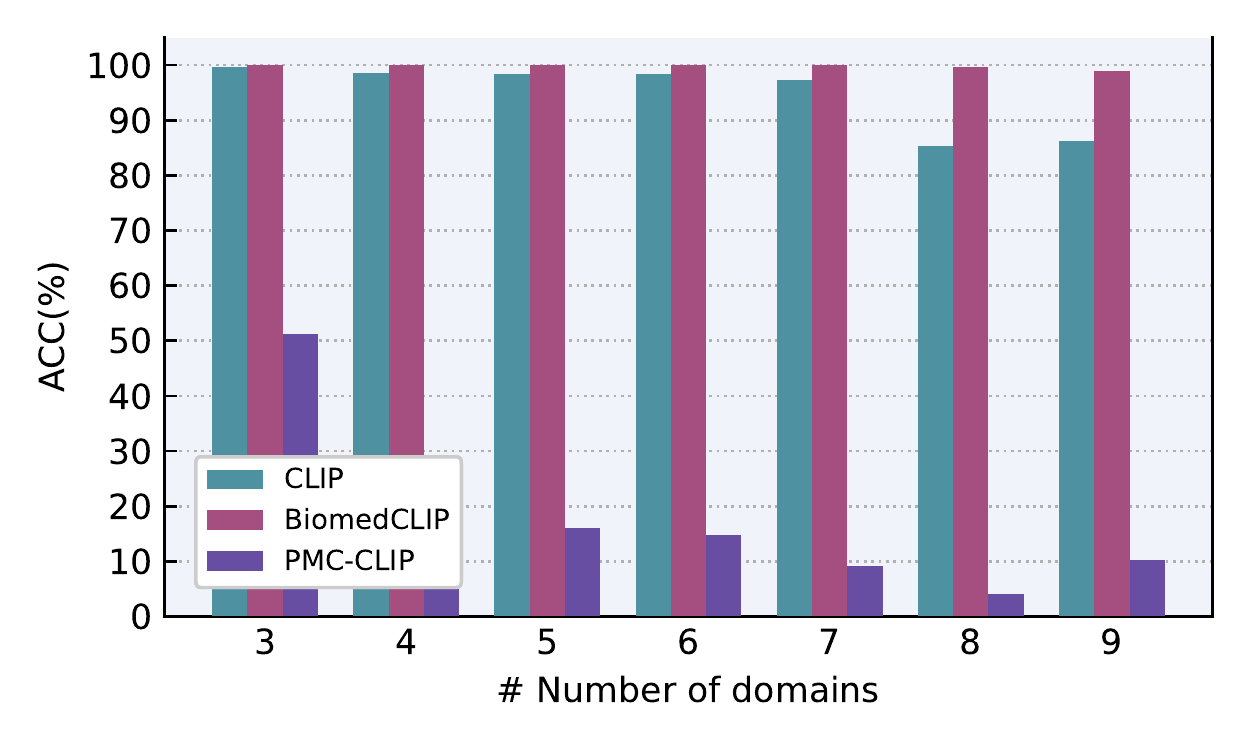}
    \caption{\textcolor{highlight}{Evaluation of domain identification using different CLIPs. 
    }}
    \label{fig:clip_cmp}
\end{figure}

\subsection{Dataset and implementations}\label{sec:dataset}
\textcolor{highlight}{
\textbf{Domain identification:}
To investigate the robustness of domain identification, which is the fundamental component of our proposed universal CAD system, we crafted a dataset covering nine medical imaging domains by randomly selecting samples from different datasets. The statistics of this dataset} \textcolor{highlight}{are presented in Table~\ref{tab:uni_stat}. 
During the evaluation, we progressively increase the number of domains following the left-to-right order in Table~\ref{tab:uni_stat} and report the accuracy (ACC) in each step.  Since the pre-trained CLIP model is utilized to perform domain identification (as shown in Fig.~\ref{fig:overview}), we also investigated the impact of different CLIPs in Sec.~\ref{sec:eval_uni}.}


\textbf{Report generation:}
For a fair comparison, the report generation performance of ChatCAD+ is assessed on the public MIMIC-CXR dataset~\cite{johnson2019mimic}.
The MIMIC-CXR is a large public dataset of chest X-ray images, associated radiology reports, and other clinical data. The dataset consists of 377,110 de-identified chest X-ray images and associated 227,827 radiology reports.
The MIMIC-CXR dataset allows us to measure the quality of diagnostic accuracy and report generation, as most report generation methods are designed specifically for chest X-ray.  The quality of report generation was tested on the official test set of the MIMIC-CXR dataset, focusing on five findings (cardiomegaly, edema, consolidation, atelectasis, and pleural effusion). Since ChatGPT was adopted as the default LLM in report generation experiments, we randomly selected 200 samples for each finding of interest due to limitations on the per-hour accesses of OpenAI API-Key, resulting in a total of 1,000 test samples.  
\textcolor{highlight}{
To evaluate the performance of report generation, we used several widely-used Natural Language Generation (NLG) metrics, including BLEU~\cite{papineni2002bleu}, METEOR~\cite{banerjee2005meteor}, and ROUGE-L~\cite{lin2004rouge}. Specifically, BLEU measures the similarity between generated and ground-truth reports by counting word overlap and we denote Corpus BLEU as C-BLEU by default. METEOR considers synonym substitution and evaluates performance on both sentence level and report level. Meanwhile, ROUGE-L evaluates the length of the longest common subsequence.
We also measured clinical efficacy (CE) of reports using the open-source CheXbert library~\cite{smit2020chexbert}. 
CheXbert takes text reports as input and generates multi-label classification labels for each report, with each label corresponding to one of the pre-defined thoracic diseases. Based on these extracted multi-hot labels, we compute precision (PR), recall (RC), and F1-score (F1) on 1,000 test samples. These metrics provide additional insight into the performance of report generation.
If not specified, we choose P3 as a default prompt design in universal interpretation and \textit{k}=3 for hierarchical in-context learning.}



\textcolor{highlight}{
\textbf{Clinical question answering:}
The evaluation of reliable interaction was performed using a subset of the CMExam dataset~\cite{liu2024benchmarking}. CMExam is a collection of multiple-choice questions with official explanations, sourced from the Chinese National Medical Licensing Examination. Since CMExam covers a diversity of areas, including many clinical-unrelated questions, we focused on a subset of the CMExam test set that specifically addressed ``Disease Diagnosis" and ``Disease Treatment," while excluding questions belonging to the department of ``Traditional Chinese Medicine". This resulted in a subset of 2190 question-answer pairs.
Since the CMExam dataset is completely in Chinese, we selected ChatGLM2~\cite{du2022glm} and ChatGLM3~\cite{du2022glm}, two 6B scale models, as the LLMs for validation. These models have demonstrated superior Chinese language capabilities compared to other open-source LLMs. We employed the official prompt design of CMExam, which required the LLM to make a choice and provide the accompanying explanation simultaneously.
We acknowledge that our proposed LLM-based knowledge retrieval involves recursive requests to the LLM and will consume a large amount of tokens, making experiments using ChatGPT and GPT-4 unaffordable for us. 
In experimental results, we report ACC and F1 for multi-choice selection, and NLG metrics for reason explanation.
}
\begin{figure*}[tbp]
    \centering
    \includegraphics[width=0.87\textwidth]{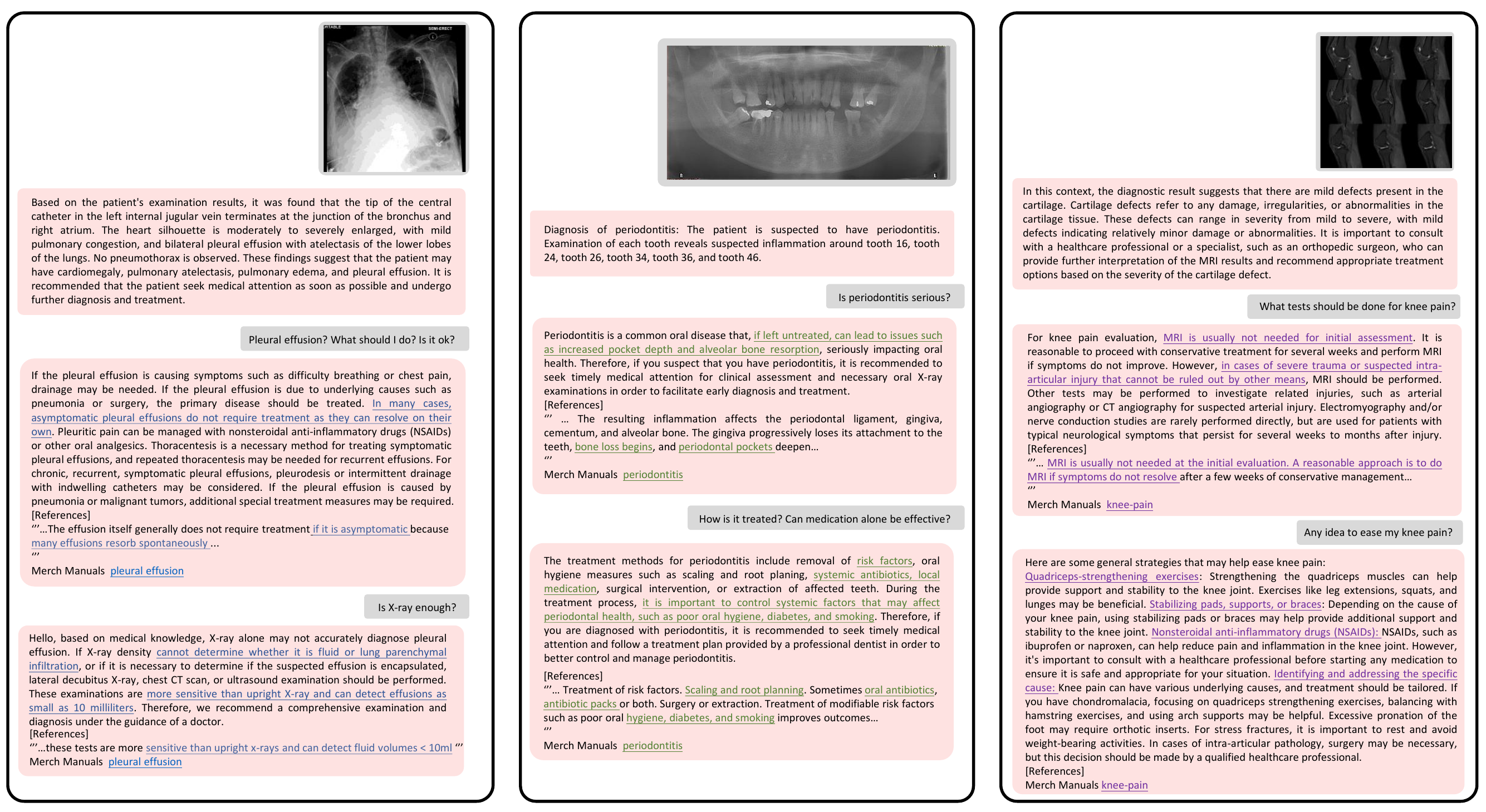}
    \caption{Examples of universal and interactive medical consultation using ChatCAD+, with ChatGPT as the default LLM. The underlined text signifies information obtained from reputable medical websites, which is not comprehensively incorporated in ChatGPT.}
    \label{interactive+}
\end{figure*}

\textbf{Implementation of domain-specific CADs:}
In Section~\ref{sec:universal}, we proposed to incorporate domain-specific CAD models to interpret images from different medical imaging domain. In this study, we adopted Chest CADs, Tooth CADs, and Knee CADs.  Chest CADs is composed of a thoracic disease classifier~\cite{ye2020weakly} and a chest X-ray report generation network~\cite{chen-acl-2021-r2gencmn}. 
The former~\cite{ye2020weakly} was trained on the official training split of the CheXpert~\cite{irvin2019chexpert} dataset, {using a learning rate (\textit{lr}) of $1e^{-4}$ for 10 epochs.} The report generation network~\cite{chen-acl-2021-r2gencmn} was trained on the training split of MIMIC-CXR~\cite{johnson2019mimic} \textcolor{highlight}{using a single NVIDIA A100 GPU. We used the Adam optimizer with a learning rate of $5e^{-5}$.}
For Tooth CAD, we utilized the periodontal diagnosis model proposed in~\cite{meilan2023hc}, which was trained on 300 panoramic dental X-ray images collected from real-world clinics with a size of 2903x1536. \textcolor{highlight}{The training was conducted on a NVIDIA A100 GPU with 80GB memory for a total of 200 epochs. We set the learning rate as $2e^{-5}$ using the Adam optimizer.}
For Knee CAD, we adopted the model proposed in~\cite{zhuang2022knee}, which was trained on 964 knee MRIs captured using a Philips Achieva 3.0T TX MRI scanner from Shanghai Sixth People's Hospital.  \textcolor{highlight}{The Adam optimizer was used with a weight decay of $1e^{-4}$.} \textcolor{highlight}{The learning rate was set as $3e^{-5}$ for Transformer and graph attention network components, and $3e^{-4}$ for others.}


\subsection{Evaluation of a Universal System}\label{sec:eval_uni}

\textcolor{highlight}{
    To investigate the effectiveness of ChatCAD+ as a universal CAD system, we evaluated it by (1) quantitatively investigating its domain identification capability when encountering images from various domains, and (2) qualitatively investigating its ability to provide reliable response in different scenarios.}
\subsubsection{Robustness of domain identification}
\textcolor{highlight}{
In Fig.~\ref{fig:clip_cmp}, we visualize the domain identification performance of different CLIP} \textcolor{highlight}{models.
Compared with the baseline CLIP model, BiomedCLIP~\cite{zhang2023large} achieves an impressive performance, showing 100\% accuracy on 2560 images from 7 different domains. It can even maintain an accuracy of 98.9\% when all 9 domains are involved, while the performance of CLIP is degraded to 86.3\%. These observations demonstrate that BiomedCLIP has an more in-depth understanding of professional medical} \textcolor{highlight}{knowledge compared with CLIP, and thus is suitable for domain identification in our proposed ChatCAD+.}

\textcolor{highlight}{
However, to our surprise, PMC-CLIP~\cite{lin2023pmc} performs rather poorly on this task. As shown in Fig.~\ref{fig:clip_cmp}, it only achieves approximate 50\% accuracy when three domains are involved. We hypothesize that this can be attributed to its lack of zero-shot capability.  While PMC-CLIP has demonstrated promising quality fof fine-tuning on donwstream tasks, its lack of zero-shot inference capability may hinder its wide-speard application.
In conclusion, BiomedCLIP can be a better choice than both CLIP and PMC-CLIP for domain identification.}

\subsubsection{Reliability of consultations}
Fig.~\ref{interactive+} demonstrates the universal application and reliability of our proposed method. The ChatCAD+ can process different medical images and provide proper diagnostic reports. Users can interact with ChatCAD+ and ask additional questions for further clarification.
By leveraging external medical knowledge, the enhanced ChatCAD+ is capable of delivering medical advice in a more professional manner. In Fig.~\ref{interactive+}, each reply from ChatCAD+ includes a reference, and the pertinent knowledge obtained from the external database is underlined for emphasis. For instance, in the scenario where the question is raised about whether a chest X-ray is sufficient for diagnosing pleural effusion, ChatGPT would simply recommend a CT scan without providing a convincing explanation. In contrast, ChatCAD+ is capable of informing the patient that CT scans have the capability to detect effusions as small as 10 milliliters, thereby providing a more detailed and informative response.

\begin{table*}[htbp]\small
      \centering
      \caption{Comparative results with previous studies and different prompt settings. P3 is the default prompt setting in this study.}
        \begin{tabular}{ll|ccc|ccc}
        \toprule
        Model & {Size} & {C-BLEU}  & {ROUGE-L} & {METEOR} & {PRE} & {REC} & {F1} \\
        \midrule
        R2GenCMN~\cite{chen-acl-2021-r2gencmn} &    244M   &   3.525       &    \textbf{18.351}   &   21.045    &  0.578    &  0.411     & 0.458 \\
        VLCI~\cite{chen2023visual}  &   357M    &   3.834     &    16.217   &    20.806    &    \textbf{0.617}    &  0.299     &  0.377\\
        ChatCAD (P3) &  256M+175B     &   3.594      &  16.791      &  23.146      &   0.526     &   0.603     & 0.553 \\
        \midrule
        ChatCAD+ (P1) &  256M+175B     &  4.073     &     16.820  &    22.700   & {0.548} & {0.484} & {0.502} \\
        ChatCAD+ (P2) &   256M+175B    &   4.407        & 17.266      &   23.302    & {0.538} & {0.597} & {0.557} \\
        ChatCAD+ (P3)&256M+175B &   \textbf{4.409}       &  17.344      &  \textbf{24.337}           &    0.531   &  \textbf{0.615}     & \textbf{0.564} \\
        \bottomrule
        \end{tabular}
      \label{tab:revised_baseline}%
\end{table*}%

\begin{table*}[htbp]\small
  \centering
  \caption{\textcolor{highlight}{Evaluation of reliable report generation using different LLMs.}}
  \resizebox{\textwidth}{!}{
    \begin{tabular}{llc|ccccccc|ccc}
    \toprule
    Model & Size  & Reliable & C-BLEU & BLEU-1 & BLEU-2 & BLEU-3 & BLEU-4 & ROUGE-L & METEOR& PRE & REC & F1 \\
    \midrule
    \multirow{2}[2]{*}{ChatGLM2~\cite{du2022glm}} & \multirow{2}[2]{*}{6B} & \XSolidBrush &2.073 & 16.999 & 3.426 & 0.926 & 0.313 & 14.119 & 23.008 &\textbf{0.518}&0.632&0.553 \\
     &       &\CheckmarkBold& \textbf{3.236} & \textbf{25.172} & \textbf{5.268} & \textbf{1.461} & \textbf{0.566} & \textbf{15.559} & \textbf{23.397}&0.501&\textbf{0.639}&\textbf{0.555} \\
     \midrule
    \multirow{2}[2]{*}{PMC-LLaMA~\cite{wu2023pmc}} & \multirow{2}[2]{*}{7B} & \XSolidBrush &\textbf{1.709} &  \textbf{15.003} & \textbf{2.432} & \textbf{0.748} & \textbf{0.313} & \textbf{13.983} & \textbf{20.590}&\textbf{0.492}&\textbf{0.702}&\textbf{0.532} \\
     &       & \CheckmarkBold  &1.059  & 6.5059 & 1.591 & {0.504} & {0.241} & {8.140} & {18.551} &0.477&{0.664}&{0.525}\\
    \midrule
    \multirow{2}[2]{*}{MedAlpaca~\cite{han2023medalpaca}} & \multirow{2}[2]{*}{7B} &\XSolidBrush& 1.554 &  30.710 & 6.143 & 1.906 & 0.819 & 12.729 & 11.865&\textbf{0.526}&0.464&0.484 \\
     &       &\CheckmarkBold&\textbf{3.031} & \textbf{30.795} & \textbf{7.107} & \textbf{2.408} & \textbf{1.107} & \textbf{14.761} & \textbf{15.711} &0.507&\textbf{0.518}&\textbf{0.505}\\
    \midrule
    \multirow{2}[2]{*}{Mistral~\cite{jiang2023mistral}} & \multirow{2}[2]{*}{7B} &\XSolidBrush& 3.261 & 27.806 & 6.068 & 1.463 & 0.458 & 16.132 & 21.999 &\textbf{0.558}&0.444&0.485\\
     &       & \CheckmarkBold & \textbf{3.547} & \textbf{30.851} & \textbf{7.178} & \textbf{1.698} & \textbf{0.493} & \textbf{17.493} & \textbf{22.347}&{0.543}&\textbf{0.597}& \textbf{0.558}\\
    \midrule
    \multirow{2}[2]{*}{LLaMA~\cite{touvron2023llama}}& \multirow{2}[2]{*}{7B}& \XSolidBrush& 0.807 & 7.575 & 1.265 & 0.337 & 0.128 & 10.016 &  17.194 &\textbf{0.495}&0.589 &0.519 \\
     &      & \CheckmarkBold& \textbf{0.966} & \textbf{7.790} & \textbf{1.525} & \textbf{0.423} & \textbf{0.178} & \textbf{10.237} & \textbf{18.344}&0.478&\textbf{0.625}&\textbf{0.535} \\
    \midrule
    \multirow{2}[2]{*}{LLaMA2~\cite{touvron2023llama2}} & \multirow{2}[2]{*}{7B} & \XSolidBrush&1.298 & 14.929 & 2.717 & 0.531 & 0.132 & 12.016 & 19.958&0.481&0.588&0.519 \\
     &    &   \CheckmarkBold & \textbf{3.579} & \textbf{24.798} & \textbf{5.826} & \textbf{1.638} & \textbf{0.693} & \textbf{16.850} & \textbf{23.950}&\textbf{0.512}&\textbf{0.561}&\textbf{0.528} \\
    \midrule
    \multirow{2}[2]{*}{LLaMA~\cite{touvron2023llama}} & \multirow{2}[2]{*}{13B}&\XSolidBrush & 0.810 & 6.824 & 1.393 & 0.331 & 0.113 & 8.959 & 16.647&{0.486}&0.452&0.459 \\
     &       &\CheckmarkBold &\textbf{0.973} & \textbf{8.295} & \textbf{1.521} & \textbf{0.439} & \textbf{0.197} & \textbf{9.813} & \textbf{18.073}&\textbf{0.498}&\textbf{0.465}&\textbf{0.463} \\
    \midrule
    \multirow{2}[2]{*}{LLaMA2~\cite{touvron2023llama2}} & \multirow{2}[2]{*}{13B}&\XSolidBrush & 2.940 & 23.536 & 5.173 & 1.390 & 0.441 & 15.053 & 21.938&0.531&0.564& 0.538\\
     &     & \CheckmarkBold & \textbf{3.650} & \textbf{ 25.080} & \textbf{6.148} & \textbf{1.728} & \textbf{0.666} & \textbf{16.646} & \textbf{24.015} &\textbf{0.533}&\textbf{0.571}&\textbf{0.543}\\
    \midrule
    \multirow{2}[2]{*}{ChatGPT~\cite{OpenAI2023ChatGPT}}& \multirow{2}[2]{*}{175B}& \XSolidBrush&3.594 &28.453 & 6.371 & 1.656 & 0.556 & 16.791 & 23.146 &0.526&0.603&0.553\\
     &&\CheckmarkBold &\textbf{4.409} & {\textbf{31.625}} & {\textbf{7.570}} & {\textbf{2.107}} & {\textbf{0.760}} &  \textbf{17.344} &   \textbf{24.337}&\textbf{0.531}&\textbf{0.615}&\textbf{0.564}\\
    \bottomrule
    \end{tabular}%
    }
  \label{tab:llms}%
\end{table*}%

\subsection{Evaluation of Report Generation}
We begin by comparing our approach to state-of-the-art (SOTA) report generation methods, while also examining the impact of various prompt designs. Subsequently, we assess the generalizability of our reliable report generation model by utilizing diverse LLMs. Following this, we explore the impact of hierarchical in-context learning by changing the number of retrieved reports. We then proceed to qualitatively showcase the advanced capabilities of ChatCAD+ in report generation.  Finally, we compare our specialist-based system against others' generalist medical AI systems.

\subsubsection{Comparison with SOTA methods}
The numerical results of different methods was compared and presented in Table~\ref{tab:revised_baseline}.
\textcolor{highlight}{
The advantage of ChatCAD+ lies in its superior performance across multiple evaluation metrics compared to both previous models and different prompt settings. Particularly, ChatCAD+ (P3) stands out with the highest scores in terms of C-BLEU, METEOR, REC, and F1.
Despite the inferiority in specific metrics, ChatCAD+ (P3) still holds a significant advantage across other metrics, which validates its advanced overall performance.
Note that ChatCAD+ (P3) shows a significant advantage over ChatCAD (P3) on all NLG and CE metrics, which underscores the rationality of hierarchical in-context learning.
These results suggest that ChatCAD+ (P3) is capable of generating high-quality responses that are not only fluent and grammatical but also clinically accurate, making it a promising model for conversational tasks.}

\textcolor{highlight}{
Additionally, the impact of prompt designs can also be observed. Generally speaking, P3 achieves the best performance while P1 leads to unsatisfying results. This phenomenon may be attributed to the similarity of P3 to human language style, as it reflects the severity of the disease using rhetoric, whereas P1 directly displays the probability without post-process.}

\subsubsection{Investigation of generalizability}
\textcolor{highlight}{
In Table~\ref{tab:llms}, we explored the generalizability of reliable report generation by adopting different LLMs.}
From the results, several noteworthy observations can be drawn:
\begin{itemize}
    \item \textcolor{highlight}{Our proposed method consistently improves the quality of generated radiology reports when combined with different LLMs, except for PMC-LLaMA. This finding demonstrates the effectiveness of our approach. }
    \textcolor{highlight}{
    \item Domain-specific LLMs, such as PMC-LLaMA and} \textcolor{highlight}{MedAlpaca, do not yield competitive enough results on reliable report generation. Despite being specifically trained on medical corpus, their performance falls short. Even with the enhancement provided by reliable report generation, MedAlpaca only achieves an F1 score of 0.505. Moreover, PMC-LLaMA does not perform well with reliable report generation, leading to significant performance degradation. We hypothesize that their inferior instruction following capability contributes to these results. Previous studies have emphasized the importance of properly controlling domain-specific fine-tuning to ensure conversational capability in LLMs~\cite{qi2024finetuning}. Unlike MedAlpaca, PMC-LLaMA fully fine-tuned its model weights during training, which could result in worse performance degradation compared to parameter-efficiently tuned MedAlpaca. We believe this is the reason why reliable report generation shows inferior performance when combined with PMC-LLaMA.}
    \item \textcolor{highlight}{A larger model size does not always guarantee better performance. For example, LLaMA (7B) and LLaMA (13B) exhibit inferior performance compared to} ChatGLM2 (6B) \textcolor{highlight}{and other 7B-scale LLMs. Thus, blindly pursuing model size may not be the optimal solution.}
    \item \textcolor{highlight}{While ChatGPT unsurprisingly achieves the best overall performance, other LLMs like LLaMA2 (13B) and Mistral (7B) still demonstrate comparable performance. In particular, Mistral (7B) achieves the second-best F1 score among all the LLMs evaluated while maintaining competent NLG metrics. This highlights the potential application of consumer-level LLMs in clinical workflows.}
\end{itemize}

\subsubsection{Ablation study of hierarchical in-context learning}
\begin{figure}[!tbp]
    \centering
    \includegraphics[width=0.42\textwidth]{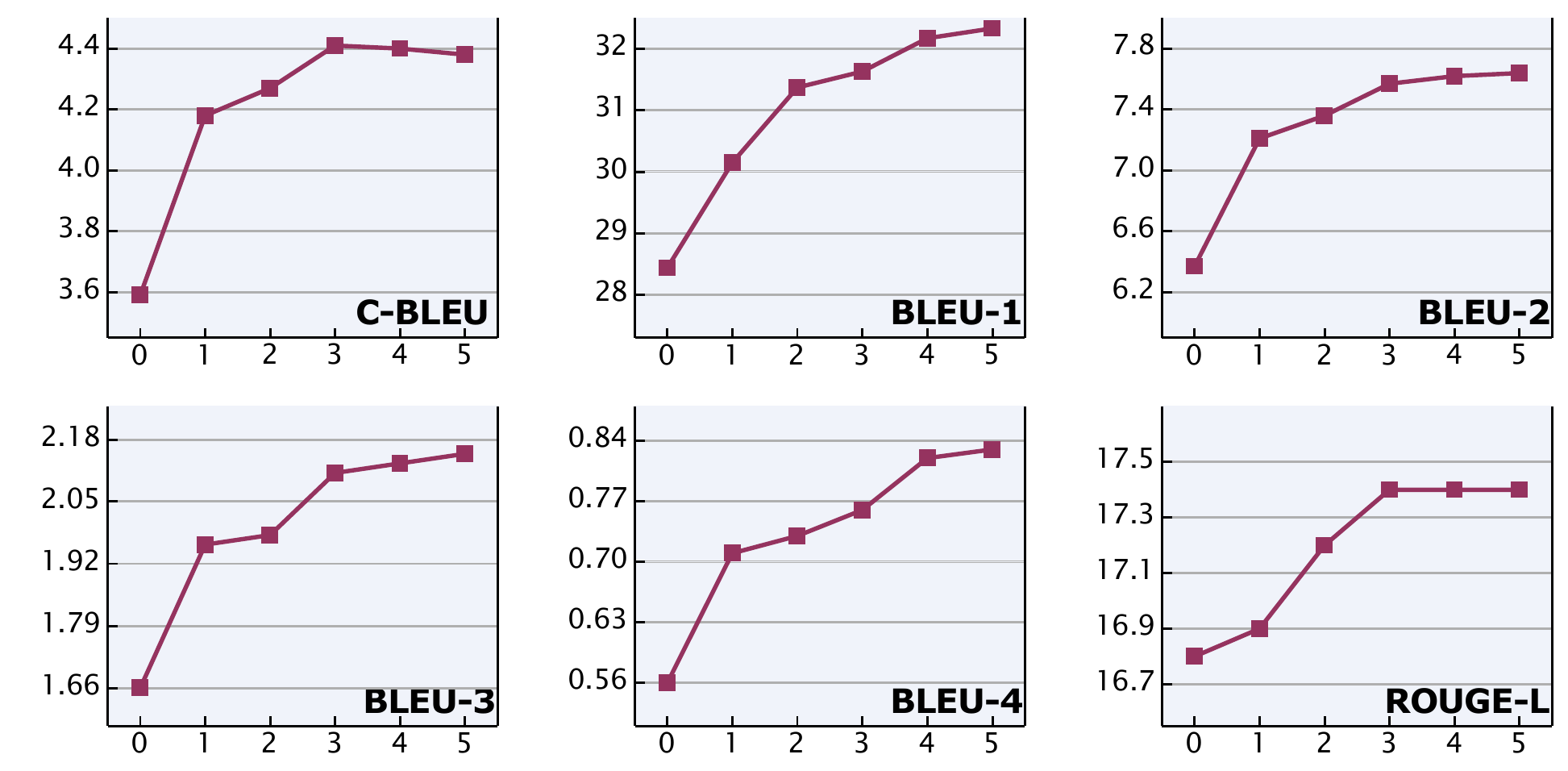}
    \caption{Ablation study of hierarchical in-context learning w.r.t. the number of retrieved reports. A range of values , varying from 0 to 5, was chosen for \textit{k}, and its influence is evaluated on several NLG metrics. }
    \label{fig:ablation_k}
\end{figure}

\begin{figure}[!tbp]
    \centering\includegraphics[width=0.42\textwidth]{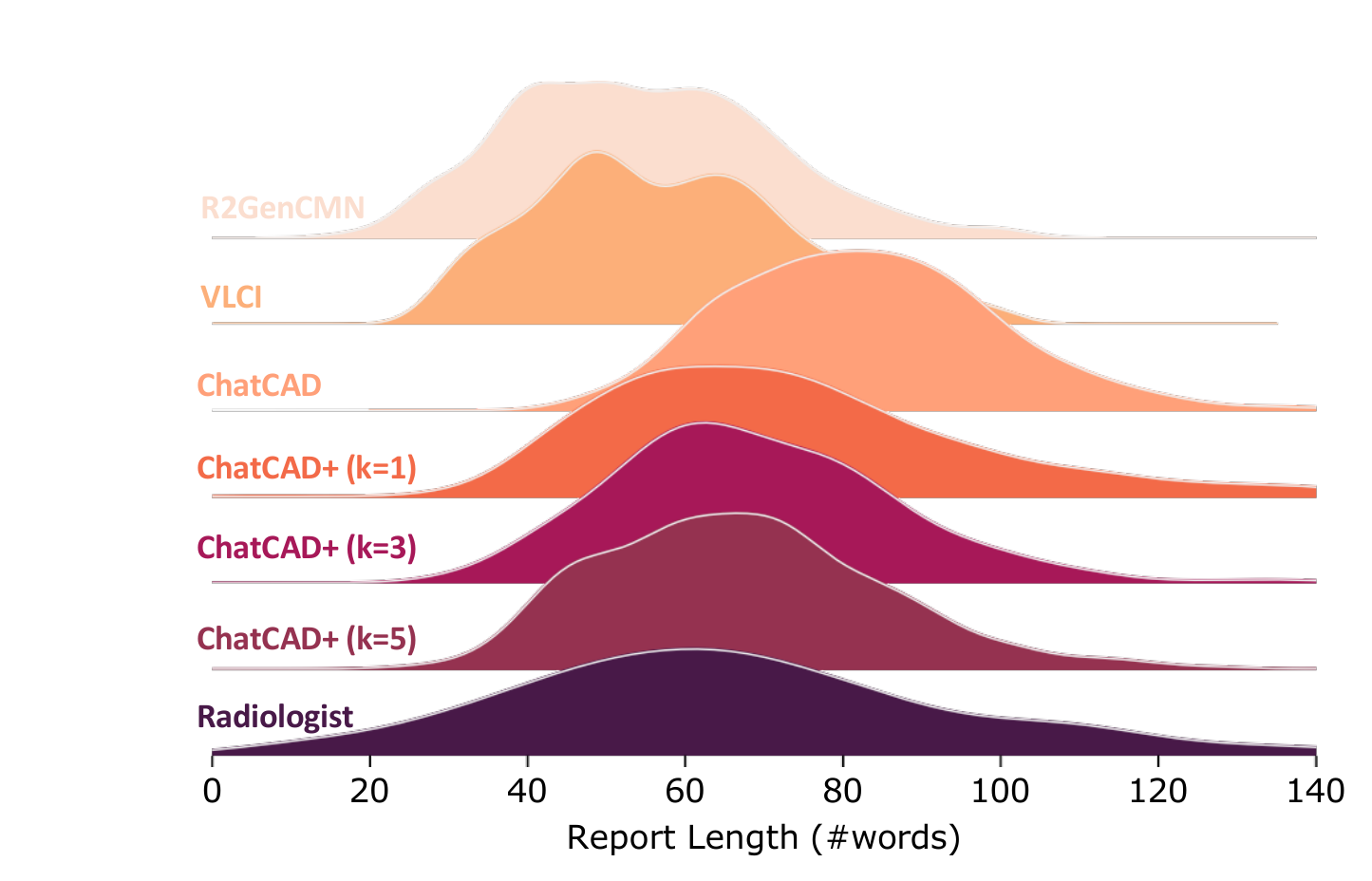}
    \caption{Distributions of report length with respect to the changing of \textit{k}.}
    \label{fig:len_model}
\end{figure}

\begin{figure*}[!tbp]
    \centering
    \includegraphics[width=0.82\textwidth]{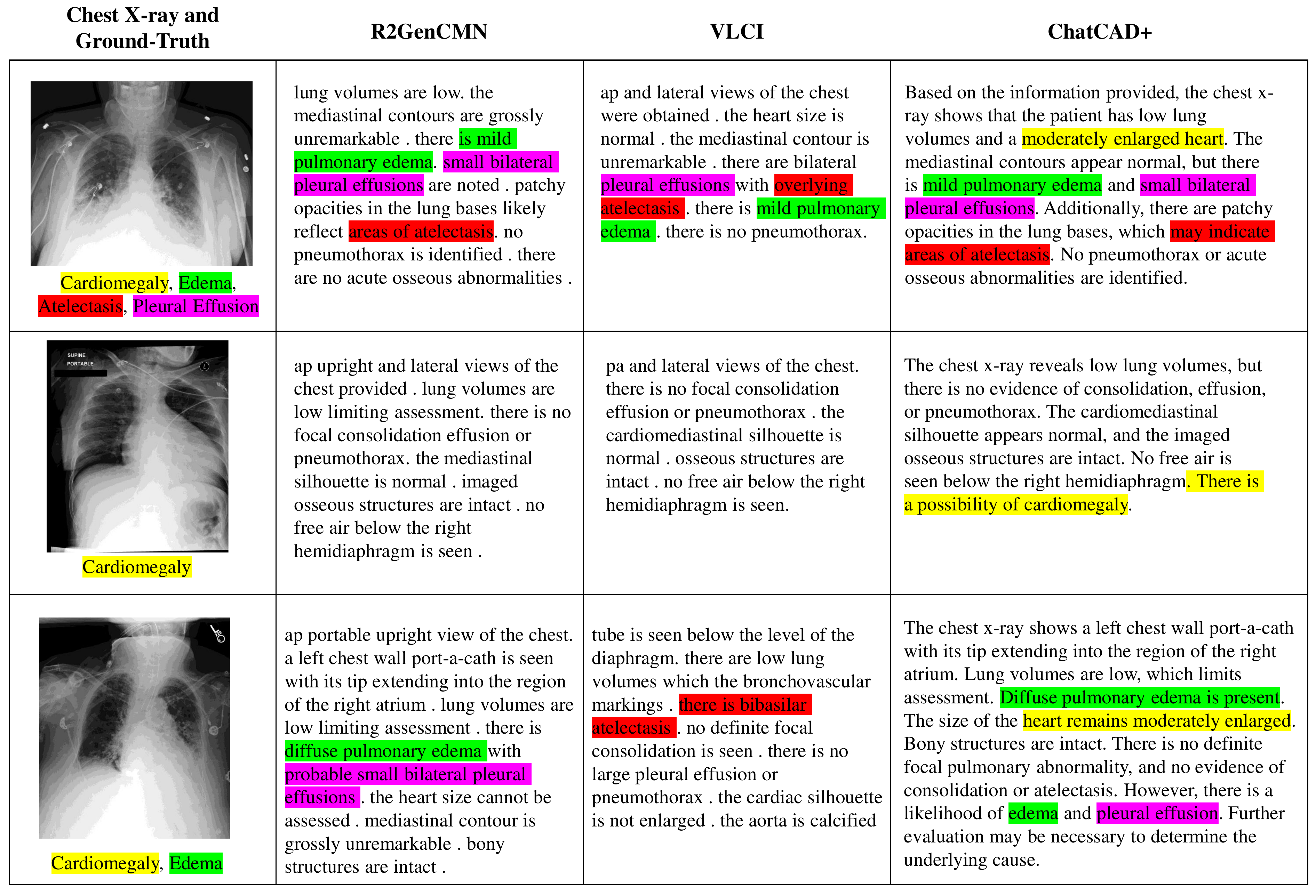}
    \caption{Qualitative analysis on different report generation methods. chest X-rays and corresponding ground-truth annotations are provided. The presence of a certain finding in a report is marked in a certain color for a clear illustration.}
    \label{fig:cmp_reports}
\end{figure*}

\begin{table}[htbp]\small
  \centering
  \caption{Performance comparison between ChatCAD+ and generalist models in the medical imaging domain. Excessively low results are represented by slash.
  }
  \resizebox{0.5\textwidth}{!}{
    \begin{tabular}{l|ccc|ccc}
    \toprule
         Method &   {C-BLEU} & {ROUGE-L} & {METEOR} & {PRE} & {REC} & {F1} \\
          \midrule
    {PMC-VQA~\cite{zhang2023pmc}}        & /  & {5.978} & {2.685} & {0.221} & {0.109} & {0.115} \\
    {RadFM~\cite{wu2023towards}}        & {/} &   {7.366} & {4.756} & {0.487} & {0.095} & {0.156} \\
    ChatCAD+ (ours)        &   \textbf{4.409}          &   \textbf{17.344}     &  \textbf{24.337}      &  \textbf{0.531}      &   \textbf{0.615}     & \textbf{0.564} \\
    \bottomrule
    \end{tabular}}%
  \label{tab:generalist}%
\end{table}%

We investigate the impact of the retrieved top-\textit{k} reports on reliable report generation in Fig.~\ref{fig:ablation_k}. As \textit{k} increases, the overall performance shows an upward trend. Moreover, in most cases, the largest improvement in performance occurs when \textit{k} changes from 0 to 1. This indicates that regardless of the quantity, as long as templates are provided as references, the quality of generated reports can be significantly enhanced. Meanwhile, the performance improvement tends to saturate around \textit{k}=3, and further increasing \textit{k} does not result in more significant improvement, which validates the rationality of the adopted default setting.

The distribution of report length is also demonstrated in Fig.~\ref{fig:len_model} by varying the value of \textit{k}. Report length serves as an important criterion\cite{chen-acl-2021-r2gencmn} to measure the similarity to  ground truth radiology reports. 
A straightforward observation is that the distribution of ChatCAD exhibits a significant shift from that of the radiologist. In contrast, ChatCAD+ shows a more fitted curve irrespective of the value of \textit{k}.

\subsubsection{Qualitative analysis}
In Fig.~\ref{fig:cmp_reports}, each row showcases a chest X-ray along with its corresponding ground-truth annotation. To enhance clarity, we have color-coded the presence of certain findings in the generated reports. This visual representation allows for a clear illustration of the differences among the methods.
For example, in the first chest X-ray image, it demonstrates the presence of four abnormalities, including cardiomegaly. However, both R2GenCMN and VLCI methods fail to acknowledge the existence of cardiomegaly in their generated reports. In the second row, the other two methods struggle to detect any abnormalities from the chest X-ray, while our method successfully identifies the possibility of cardiomegaly. The same trend can be observed in the third row.
By including this qualitative analysis, we aim to highlight the advantages of our ChatCAD+ method and provide a comprehensive evaluation of the template retrieval system (hierarchical in-context learning). This comparison demonstrates the improved report quality achieved by our proposed method.

\subsubsection{Comparison with generalists}
While ChatCAD+ aims to integrate multiple specialist AI models incrementally, some recent studies~\cite{wu2023pmc,wu2023towards} have investigated generalist AI models for medical imaging. Hence, we quantitatively compare ChatCAD+ to generalists to investigate its superiority.

\textcolor{highlight}{
Existing generalist models typically follow the architecture of DeepMind's Flamingo~\cite{alayrac2022flamingo} and Salesforce's BLIP-2~\cite{li2023blip}. For instance, RadFM~\cite{wu2023towards} and PMC-VQA~\cite{zhang2023pmc} are trained based on Flamingo and BLIP-2,  respectively. }
\textcolor{highlight}{
We compared ChatCAD+ with PMC-VQA and Flamingo using the same test set, and conducted zero-shot inference, as shown in Table~\ref{tab:generalist}. Overall, our method demonstrates significant advantages over the generalist models. It is worth noting that these generalist models exhibit extremely low values on C-BLEU, indicating} \textcolor{highlight}{that generalist models can only generate very short responses and lack the ability to provide comprehensive diagnostic information like ours.
Generally speaking, a generalist model implies that a single image encoder needs to interpret semantic information for all possible input images. This is challenging to achieve in the medical imaging domain due to distinct morphological and clinical semantic differences among different medical imaging modalities, which require a deep understanding of professional medical knowledge. Based on these reasons, we believe that it is still not the ChatGPT time for medical imaging analysis, and specialist models will long been a preferable solution. Such preference of specialist model is also mentioned in a related survey~\cite{zhang2023challenges}.
}

\textcolor{highlight}{
It is also important to note that a major advantage of ChatCAD+ lies in their capability for continual learning. 
By incorporating the LLM, we can combine decisions from multiple CAD models, enabling fine-tuning of each CAD model individually and also their incremental ensembling. 
This}  \textcolor{highlight}{flexibility allows us to adapt and integrate new models rapidly in response to emerging situations without the need of huge computation resources. 
}


\subsection{Evaluation of clinical question answering}

\begin{table}[tbp]
  \centering
  \caption{Validation of reliable interaction from perspectives of both multi-choice selection and reason explanation.}
    \begin{tabular}{lcccc}
    \toprule
          \multirow{2}[2]{*}{Method}& \multicolumn{2}{c}{Multi-choice}  & \multicolumn{2}{c}{Reason Explanation} \\
          \cmidrule(lr){2-3} \cmidrule(lr){4-5}
           & ACC   & F1   & C-BLEU  &   ROUGE-L \\
    \midrule
    ChatGLM2~\cite{du2022glm} & 0.435 & 0.427 & 2.430    & 11.396 \\
    ChatGLM2+Ours & \textbf{0.465} & \textbf{0.461} & \textbf{7.930}   & \textbf{18.957} \\
    \midrule
    ChatGLM3~\cite{du2022glm} & 0.458 & 0.455 & 3.270     & 14.915 \\
    ChatGLM3+Ours & \textbf{0.472} & \textbf{0.505} &     \textbf{7.974}   &  \textbf{19.302}\\
    \bottomrule
    \end{tabular}
  \label{tab:interactionQA}%
\end{table}%

\begin{table*}
\centering
\resizebox{0.85\textwidth}{!}{
    \begin{tabular}{|p{\textwidth}|}
        \hline
    \textbf{Q1} Is pleural effusion serious? \\
    \hline
    \textbf{K1 (Baseline)} 
    Under normal circumstances, a thin layer of pleural fluid (10-20mL) covers the surfaces of the visceral and parietal pleura, with a composition similar to plasma but with lower protein content (\textless1.5g/dL [\textless15 g/L]). 
     \tpurp{Pleural effusions are usually classified into exudates, transudates, and fluid accumulation based on laboratory characteristics of the fluid (see criteria for diagnosing exudative pleural effusions in the table).Regardless of unilateral or bilateral involvement, leaky effusions usually require little investigation and can be treated, while exudative effusions require identification of the underlying cause. Various factors can cause pleural effusions (see the table for causes of pleural effusions). Leaky pleural effusions are caused by an increase in systemic circulatory hydrostatic pressure and a decrease in plasma colloid osmotic pressure.}  \\
    \hline
    \textbf{K1 (Ours)} 
\tgrn{Some pleural effusions are asymptomatic and are discovered incidentally during physical examination or on chest x-ray. Many cause dyspnea, pleuritic chest pain, or both. Pleuritic chest pain, a vague discomfort or sharp pain that worsens during inspiration, indicates inflammation of the parietal pleura. Pain is usually felt over the inflamed site, but referred pain is possible.} The posterior and peripheral portions of the diaphragmatic pleura are supplied by the lower 6 intercostal nerves, and irritation there may cause pain in the lower chest wall or abdomen that may simulate intra-abdominal disease. Irritation of the central portion of the diaphragmatic pleura, innervated by the phrenic nerves, causes pain referred to the neck and shoulder.
\tgrn{Physical examination reveals absent tactile fremitus, dullness to percussion, and decreased breath sounds on the side of the effusion. These findings can also be caused by pleural thickening.}

... 
Pericardial rub is best heard over the left border of the sternum in the 3rd and 4th intercostal spaces, is characteristically a to-and-fro sound synchronous with the heartbeat, and is not influenced significantly by respiration. Sensitivity and specificity of the physical examination for detecting effusion are probably low.\\
    \hline
    \textbf{A1 (ChatGPT Only)} 
    Pleural effusion can be serious depending on the underlying cause and the amount of fluid accumulated in the pleural space. It can be a symptom of various medical conditions such as congestive heart failure, pneumonia, cancer, or kidney disease. If left untreated, pleural effusion can lead to breathing difficulties and complications. It's important to consult with a healthcare provider for proper diagnosis and treatment. \\
    \hline
    \textbf{A1 (Ours)} 
    \tgrn{The severity of pleural effusion needs to be determined based on the patient's specific situation. Generally, a large amount of effusion can cause symptoms such as dyspnea and chest pain, but some patients may not have obvious symptoms. Physical examination can reveal reduced breath sounds on the effusion side, dullness to percussion, and possibly pleural friction rub.} However, the sensitivity and specificity of physical examination for pleural effusion are both low, so clinical symptoms and imaging examinations need to be combined to determine the severity of the effusion and treatment plan. Common causes of pleural effusion include heart failure, pneumonia, and tuberculosis \\
    \hline
    
    \hline
    \textbf{Q2} What should I do after getting lung atelectasis?\\
    \hline
    \textbf{K2 (Baseline)} do not recommend oral or intravenous corticosteroids because they have not shown efficacy in metal fume fever. Patients with pre-existing lung disease (such as asthma or chronic obstructive pulmonary disease) rarely require treatment and hospitalization due to worsening of pre-existing conditions. If severe cases of either syndrome cause respiratory distress, treatment methods used for ARDS (such as mechanical ventilation and positive end-expiratory pressure [PEEP]) may also be used. "  , "Atelectasis":  "Abs": "\tpurp{Atelectasis is collapse of lung tissue with loss of volume. Patients may have dyspnea or respiratory failure if atelectasis is extensive. They may also develop pneumonia. Atelectasis is usually asymptomatic, but hypoxemia and pleuritic chest pain may be present in certain cases. Diagnosis is by chest x-ray. Treatment includes maintaining coughing and deep breathing and treating the cause."} \\
    \hline
    \textbf{K2 (Ours)} 
    \tgrn{Maximizing cough and deep breathing
    If obstruction by tumor or foreign body is suspected, bronchoscopy
Evidence for the efficacy of most treatments for atelectasis is weak or absent. Nonetheless, commonly recommended measures include chest physiotherapy to help maintain ventilation and clearance of secretions, and encouragement of lung expansion techniques such as directed cough, deep breathing exercises, and use of an incentive spirometer. In ambulatory patients, exercise (eg, walking) is a desirable way to promote deep breathing.}
For patients who are not intubated and do not have excessive secretions, continuous positive airway pressure may help. For patients who are intubated and mechanically ventilated, positive end-expiratory pressure and/or higher tidal volume ventilation may help.

...
Most importantly, the cause of atelectasis (eg, mucous plug, foreign body, tumor, mass, pulmonary effusion) should be treated. For persistent mucous plugging, nebulized dornase alfa and sometimes bronchodilators are tried. N-Acetylcysteine is usually avoided because it can cause bronchoconstriction. 
If other measures are ineffective or if a cause of obstruction other than mucous plugging is suspected, bronchoscopy should be done.\\
    \hline
    \textbf{A2 (ChatGPT Only)}
    I'm not a doctor, but I can provide you with general information on lung atelectasis. It's important to follow your healthcare provider's advice for your specific situation. In general, treatment for lung atelectasis may involve addressing the underlying cause, such as removing any blockages in the airways, treating infections, or providing oxygen therapy. Your healthcare provider may also recommend techniques to help re-expand the collapsed lung tissue, such as \tgrn{deep breathing exercises, chest physiotherapy, or using a device called a spirometer}. Make sure to follow up with your healthcare provider for proper diagnosis and treatment.\\
    \hline
    \textbf{A2 (Ours)} 
    \tgrn{The treatment methods for pulmonary atelectasis include maintaining cough and deep breathing, chest physical therapy, exercise, continuous positive airway pressure, etc.} However, the evidence for the effectiveness of these treatments is not sufficient, so the treatment plan needs to be developed according to the specific situation of the patient. If there is suspicion of tumor or foreign body obstruction, bronchoscopy examination is needed. \tgrn{Opioid drugs can be used to relieve severe pleural pain, but cough suppressant drugs should be avoided.} For the causes of pulmonary atelectasis such as mucus plugs, foreign bodies, tumors, masses, pleural effusions, etc., targeted treatments should be used.\\
    \hline
    \end{tabular}}
    \caption{Comparison between knowledge retrieval methods. ChatGPT is selected as the default LLM. Completely unrelated knowledge is marked in \tred{red}, partially related knowledge is  marked in \tpurp{purple}. Related and important knowledge is in \tgrn{green}.}
    \label{tab:example12}
\end{table*}

\begin{table*}
\centering
\resizebox{0.85\textwidth}{!}{
  \begin{tabular}{|p{\textwidth}|}
    \hline
    \textbf{Q3} How to treat periodontitis? Is taking medication enough? \\
    \hline
    \textbf{{K3 (Baseline)}} \tred{Esophageal cancer can occur at a young age. Unna-Thost disease and Vorner disease: autosomal dominant inheritance. Papillon-Lefèvre syndrome: autosomal recessive inheritance, onset within 6 months after birth. Severe periodontitis can lead to tooth loss. Vohwinkel syndrome: autosomal dominant inheritance, patients may experience progressive toe and finger amputation and high-frequency hearing loss. Diffuse non-epidermolytic palmoplantar keratoderma: this autosomal dominant inherited form develops in infancy, causing a sharply defined symmetric keratoderma affecting the entire palm and sole. Treatment: Symptomatic treatment includes the use of emollients, keratolytic agents, and physical methods for removal.} \\
    \hline
    \textbf{K3 (Ours)} 
    Treatment of risk factors
    Scaling and root planing
    \tgrn{Sometimes oral antibiotics, antibiotic packs, or both}
    Surgery or extraction
\tgrn{Treatment of modifiable risk factors such as poor oral hygiene, diabetes, and smoking improves outcomes.} 
For all forms of periodontitis, the first phase of treatment consists of thorough scaling (professional cleaning with hand or ultrasonic instruments) and root planing (removal of diseased or toxin-affected cementum and dentin followed by smoothing of the root) to remove plaque and calculus deposits. Thorough home oral hygiene is necessary and includes careful brushing and flossing to help clean. It may include chlorhexidine swabs or rinses. A therapist should help teach the patient how to do these procedures. The patient is re-evaluated after 3 weeks. If pockets are no deeper than 4 mm at this point, the only treatment needed is regular cleanings. Sometimes a flap of gum tissue is made to allow access for scaling and planing of deeper parts of the root.
\tgrn{If deeper pockets persist, systemic antibiotics can be used. A common regimen is amoxicillin 500 mg orally 3 times a day for 10 days.} In addition, a gel containing doxycycline or microspheres of minocycline can be placed into isolated recalcitrant pockets. These drugs are resorbed in 2 weeks.
Another approach is to surgically eliminate the pocket and recontour the bone (pocket reduction/elimination surgery) so that the patient can clean the depth of the normal crevice (sulcus) between the tooth and gingiva. In certain patients, regenerative surgery and bone grafting are done to encourage alveolar bone growth. Splinting of loose teeth and selective reshaping of tooth surfaces to eliminate traumatic occlusion may be necessary. \tgrn{Extractions are often necessary in advanced disease.} Contributing systemic factors should be controlled before initiating periodontal therapy.
Ninety percent of patients with necrotizing ulcerative periodontitis due to HIV (HIV-associated periodontitis) respond to combined treatment with scaling and planing, irrigation of the sulcus with povidone-iodine (which the dentist applies with a syringe), regular use of chlorhexidine mouth rinses, and systemic antibiotics, usually metronidazole 250 mg orally 3 times a day for 14 days.
Localized aggressive periodontitis requires periodontal surgery plus oral antibiotics (eg, amoxicillin 500 mg 4 times a day or metronidazole 250 mg 3 times a day for 14 days). \\
    \hline
    \textbf{A3 (ChatGPT Only)} Treating periodontitis usually involves a combination of professional dental treatments. Medication alone may not be enough to fully treat periodontitis. 
    Professional dental treatments may include scaling and root planing to remove plaque and tartar from the teeth and roots, as well as possibly surgical procedures to correct any damage to the gums and bone.
    It's important to follow your dentist's recommendations closely and attend regular dental check-ups to monitor your progress and make any necessary adjustments to your treatment plan.\\
    \hline
    \textbf{A3 (Ours)} The treatment of periodontitis requires a comprehensive approach that considers risk factors, scaling and root planing, oral antibiotics, and local medications. For all cases of periodontitis, the first stage of treatment involves thorough cleaning, scaling and root planing, and reinforcement of daily oral hygiene. \tgrn{If deep periodontal pockets are present, systemic antibiotics may be used, with amoxicillin 500mg orally, three times a day for 10 days being a commonly used method.} The second stage of treatment involves surgical removal of periodontal pockets and bone reshaping to allow the patient to self-clean the base of the gingival sulcus. Localized aggressive periodontitis requires periodontal surgery in combination with oral antibiotics. At the same time, systemic factors such as poor oral hygiene, diabetes, and smoking must be controlled to improve the prognosis. Therefore, drug therapy alone is not sufficient and comprehensive treatment is necessary. \\
    \hline

    \hline
    \textbf{Q4} What tests should be done for knee pain? \\
    \hline
    \textbf{{K4 (Baseline)}} \tpurp{Patellar tendon tear: the patella is palpable and displaced upwards (high-riding patella). Transverse patellar fracture: a gap between the two pieces of the patella may be palpable. However, due to significant local swelling, these symptoms may be mistaken for knee ligament injury with joint effusion. If a patient presents with knee swelling and pain after injury, the clinician will ask the patient to sit down and attempt to extend the injured leg to test knee extension actively or lie flat and elevate the injured leg.} After knee swelling following trauma,  remember to test the knee's ability for independent extension. Routine knee joint X-ray examination should be performed. \tpurp{Knee X-rays can show a high or low position of the patella. X-rays can reveal dislocation or patellar fractures, but they may also appear normal.} \\
    \hline
    \textbf{K4 (Ours)} 
Not every patient requires x-rays. However, anteroposterior, lateral, and oblique x-rays are often taken to exclude fractures. The Ottawa knee rules are used to limit x-rays to patients more likely to have a fracture that requires specific treatments. X-rays should be taken only if one of the following is present:
    Age $>$ 55,
    Isolated tenderness of the patella (with no other bone tenderness of the knee),
    Tenderness of fibular head,
    Inability to flex the knee 90°,
    Inability to bear weight both immediately and in the emergency department for 4 steps (with or without limping).\\
\tgrn{MRI is usually not needed at the initial evaluation. A reasonable approach is to do MRI if symptoms do not resolve after a few weeks of conservative management. However, MRI is often done when severe injury or significant intra-articular injury is suspected or cannot be otherwise excluded.
Other tests may be done to check for related injuries:
    Arteriography or CT angiography to check for suspected arterial injuries
    Electromyography and/or nerve conduction studies, which are rarely done immediately but more typically done when nerve symptoms persist weeks to months after the injury.}\\
    \hline
    \textbf{A4 (ChatGPT Only)} When diagnosing knee pain, there are several tests that healthcare professionals may recommend to determine the underlying cause. These tests may include:
    (1) Physical examination: A healthcare provider will assess your knee by examining its range of motion, stability, and tenderness.
    (2) X-rays: X-rays can help identify issues such as fractures, arthritis, or bone abnormalities in the knee.
    (3) MRI (Magnetic Resonance Imaging): MRI scans provide detailed images of soft tissues like ligaments, tendons, and cartilage, which can help diagnose conditions such as ligament tears or cartilage damage.
    (4) CT scan (Computed Tomography): CT scans can provide detailed images of the bones and joints in the knee, helping to diagnose fractures or bone abnormalities.
    (5) Ultrasound: Ultrasound imaging can help evaluate soft tissue structures like tendons and ligaments for injuries or inflammation.
    (6) Arthroscopy: In some cases, a minimally invasive procedure called arthroscopy may be recommended to directly visualize and diagnose issues within the knee joint.
    It's important to consult with a healthcare professional to determine which tests are necessary based on your specific symptoms and medical history.\\
    \hline
    \textbf{A4 (Ours)} For knee pain evaluation, \tgrn{MRI is usually not needed for initial assessment.} \tgrn{It is reasonable to proceed with conservative treatment for several weeks and perform MRI if symptoms do not improve.} However, in cases of severe trauma or suspected intra-articular injury that cannot be ruled out by other means, MRI should be performed. Other tests may be performed to investigate related injuries, such as arterial angiography or CT angiography for suspected arterial injury. Electromyography and/or nerve conduction studies are rarely performed directly, but are used for patients with typical neurological symptoms that persist for several weeks to months after injury. Therefore, which tests are needed depends on the specific situation of the patient. \\
    \hline
  \end{tabular}}
  \caption{Comparison between knowledge retrieval methods. ChatGPT is selected as the default LLM. Completely unrelated knowledge is marked in \tred{red}, partially related knowledge is  marked in \tpurp{purple}. Related and important knowledge is in \tgrn{green}.}
  \label{tab:example34}
\end{table*}

To evaluate the efficacy of clinical question answering, we adopt a comprehensive approach that combines quantitative analysis (through testing on the CMExam dataset) and qualitative analysis (by posing potential questions in real-world scenarios). This enables a thorough examination and provides insight into how the system functions in practical context.

\textcolor{highlight}{
In Table~\ref{tab:interactionQA}, we choose ChatGLM2 and ChatGLM3 as the LLM and present the performance of multi-choice selection and reason explanation with and without our proposed knowledge retrieval method on a subset of CMExam dataset (referring to Sec.~\ref{sec:dataset}). It is important to note that the results of multi-choice selection are directly generated by the LLM via prompt design, without intervention of any text encoders. We observe that our proposed method can consistently improve the performance of question answering for both ChatGLM2 and ChatGLM3.
Interestingly, we find that the generation of reason explanations is significantly enhanced compared to multi-choice selection. For ChatGLM2, the ROUGE-L score is improved from 11.396 to 18.957, and for ChatGLM3, it is improved from 14.925 to 19.302. However, the improvement in multi-choice selection is not pronounced. We hypothesize that this discrepancy arises because the LLM may occasionally select correct answer based on unrigorous or even incorrect reasoning. In such cases, the LLM's reliability is compromised. However, with our proposed method, the reasoning behind the correct answer becomes more convincing, enabling reliable interaction for patients in need.}

Considering qualitative analysis, the effectiveness of our proposed LLM-based knowledge retrieval method is compared with a baseline method that does not rely on a LLM. Such methods tend to follow the paradigm of LangChain~\cite{langchain-glm}, which involves dividing the text into paragraphs and then utilizing sentence transformers to compare the similarity between the user query and all paragraphs.  This kind of knowledge retrieval method can hardly handle the professional medical knowledge database. We select the implementation in~\cite{langchain-glm} as the baseline and compare it with our proposed method in Table~\ref{tab:example12} and Table~\ref{tab:example34}, where ``K'' and ``A'' indicate ``retrieved knowledge” and ``answer from the LLM”, respectively. Note that ChatGPT is the default LLM in this qualitative analysis.

\textcolor{highlight}{
    In accordance with the structure of the Merck Manual, each medical topic is divided into five main sections: ``abstract", ``symptoms and signs", ``diagnosis", ``treatment" and ``prognosis" as shown in Fig.~\ref{fig:interaction_overview}. Therefore, to assess the effectiveness of knowledge retrieval, all the questions presented in Table~\ref{tab:example12} and Table~\ref{tab:example34} are implicitly associated with certain sections.}
\textcolor{highlight}{
If the retrieved knowledge comes from the corresponding section, we label it as ``\tgrn{related}". If the knowledge is retrieved from a different section but still belongs to the same medical topic, we label it as ``\tpurp{partially related}". Otherwise, if the knowledge comes from a different medical topic altogether, we consider it ``\tred{unrelated}".}
\textcolor{highlight}{
For example, in the case of Q1 (``Is pleural effusion serious?"), the target section is ``symptoms and signs''. Unfortunately, the baseline method retrieves information from the ``abstract'' section, which is only ``\tpurp{partially related}'' and cannot directly provide a reliable answer.}
\textcolor{highlight}{    
In the case of Q3, we ask for the treatment of periodontitis. Our proposed method successfully retrieves the ``treatment'' section of periodontitis, while the baseline method mistakenly retrieves information about palmoplantar keratodermas. 
As a result, we labeled it as ``\tred{unrelated}".
We hypothesize the reason behind this mistake is that palmoplantar keratodermas can lead to secondary infections of periodontitis. }
\textcolor{highlight}{  
These results clearly illustrate the advancement of our method. On the contrary, ~\cite{langchain-glm} struggles to identify relevant medical topics due to limited vocabulary size and understanding. This issue can be worse if the question of the user does not explicitly point to any medical topics or involve multiple medical entities.}

\textcolor{highlight}{  
We also demonstrate the answer of ChatGPT without additional knowledge. Results reveal that ChatGPT's responses are less useful and accurate. Specifically, for the first question, ChatGPT fails to elaborate on the fact that the severity of pleural effusion depends on the specific circumstances, instead merely stating that untreated pleural effusion can lead to complications. This omission prevents a thorough understanding of the condition.
Furthermore, in Q4, ChatGPT provides an extensive list of possible tests for the patient, which can potentially cause unnecessary anxiety and expenses. In contrast, our reliable interaction highlights that an MRI is not imperative for initial assessment and should only be considered if knee pain does not improve after several weeks of conservative treatment. This clarification offers valuable guidance and prevents unwarranted medical interventions. We believe that this comparison could further highlight the effectiveness of our proposed reliable interaction.}

\section{Conclusion}
In this work, we have developed ChatCAD+, a universal and reliable interactive CAD system that can accept images of different domains as valid input. We expect that the proposed paradigm of universal CAD networks can motivate the advancement of cross-domain model literature in medical image domain.
Furthermore, the hierarchical in-context learning enhances the report generation ability of ChatCAD+ in a training-free manner and let it act more like a radiologist. The LLM-based knowledge retrieval method enables the LLM to search by itself, and utilize the external medical knowledge database to provide convincing medical treatments.
In this way, our ChatCAD+ has utilized the Merck Manuals database as its primary external knowledge database, which may be considered limited in its scope. However, we are confident in the versatility of our proposed LLM-based knowledge retrieval method, as it can be readily adapted to incorporate other databases. By doing so, we anticipate that the reliability of ChatCAD+ will be greatly enhanced.
There are several limitations that must be acknowledged. First, while we have succeeded in enhancing the report generation capacity of our model to closely resemble that of a radiologist, it relies on additional report database.
Second, the knowledge retrieval method we employed is relatively slow, as its efficacy is closely related to the response speed of the OpenAI API. These limitations provide an opportunity for future research to address them and improve upon the current findings.


\bibliography{ref.bib}
\bibliographystyle{IEEEtran}
\end{document}